\documentclass[lettersize,11pt,onecolumn, draftclsnofoot]{IEEEtran}

\usepackage{geometry}
\usepackage{wrapfig}

\usepackage{color, colortbl}

\usepackage{amsmath, amsthm, amssymb, float, breakcites, enumitem}

\usepackage{multirow} 
\usepackage{mathtools}
\usepackage{algorithm,algpseudocode}

\usepackage{pifont}

\usepackage{blindtext}
\usepackage{lipsum}

\usepackage{multirow}
\usepackage{graphicx}
\usepackage{listings}

\usepackage{bbm}

\usepackage[most]{tcolorbox}
\usepackage{adjustbox}
\usepackage{booktabs}
\usepackage{hyperref}
\usepackage{url}
\usepackage{cite}
\usepackage[para,online,flushleft]{threeparttable}

\usepackage{todo}

\usepackage{fontawesome}

\usepackage{amsmath,amsfonts,bm}

\def\eqref#1{(\ref{#1})}

\def\1{\bm{1}}

\DeclareMathAlphabet{\mathsfit}{\encodingdefault}{\sfdefault}{m}{sl}
\SetMathAlphabet{\mathsfit}{bold}{\encodingdefault}{\sfdefault}{bx}{n}

\newcommand{\E}{\mathbb{E}}

\DeclareMathOperator*{\argmin}{arg\,min}

\DeclareMathOperator*{\minimize}{\text{minimize}}
\DeclareMathOperator*{\maximize}{\text{maximize}}

\DeclareMathOperator*{\st}{\text{subject to}}
\DeclareMathAlphabet\mathbfcal{OMS}{cmsy}{b}{n}
\newcommand{\Def}[0]{\mathrel{\mathop:}=}

\newcommand{\by}{\mathbf{y}}
\newcommand{\BZ}{\mathbf{Z}}
\newcommand{\BA}{\mathbf{A}}
\newcommand{\BB}{\mathbf{B}}

\newcommand{\bc}{\mathbf{c}}

\newcommand{\bd}{\mathbf{d}}
\newcommand{\bdelta}{\boldsymbol{\delta}}

\newcommand{\btheta}{\boldsymbol{\theta}}

\newcommand{\bphi}{\boldsymbol \phi}

\usepackage{algorithm,algpseudocode}
\algnewcommand{\algorithmicforeach}{\textbf{for each}}
\algdef{SE}[FOR]{ForEach}{EndForEach}[1]
  {\algorithmicforeach\ #1\ \algorithmicdo}
  {\algorithmicend\ \algorithmicforeach}

\usepackage{pifont}

\usepackage{color, colortbl}
\definecolor{Gray}{gray}{0.93}
\definecolor{Orange}{rgb}{1,0.5,0}
\definecolor{DGray}{gray}{0.83}
\definecolor{LightCyan}{rgb}{0.88,1,1}

\definecolor{Gray}{gray}{0.93}

\definecolor{light-gray}{gray}{0.8}

\newcommand{\IG}{{\text{IG}}}

\newcommand{\task}{\mathcal{T}}
\newcommand{\din}{\mathcal D^{\texttt{tr}}}
\newcommand{\dout}{\mathcal D^{\texttt{val}}}

\newcommand{\BBLO}{{\text{LU-BLO}}}
\newcommand{\CBLO}{{\text{LC-BLO}}}
\newcommand{\GU}{{\text{GU}}}
\newcommand{\IF}{{\text{IF}}}
\newcommand{\VF}{{\text{VF}}}

\newcommand{\recommend}{\faThumbsOUp}

\newtcolorbox{mybox}[3][]
{
  colframe = #2!35,
  colback  = #2!5,
  coltitle = #2!20!black,  
  title    = {#3},
  #1,
}

\newtcolorbox{examplebox}[2][]
{
  colback=green!10,
  colframe=green!50!black,
  coltitle = white,  
  title    = {#2},
  #1,
}

\newtcolorbox{algbox}[2][]
{
  colframe = orange!55!yellow!40,
  colback  = yellow!5,
  coltitle = black,  
  title    = {#2},
  #1,
}

\begin{document}
\title{An Introduction to Bi-level Optimization: Foundations and  Applications in Signal Processing and Machine Learning}

\author{Yihua Zhang$^{1}$, ~ Prashant Khanduri$^{2}$, 
~ Ioannis Tsaknakis$^3$,
~ Yuguang Yao$^1$, \\
~ Mingyi Hong$^3$, ~ Sijia Liu$^{1,4}$
\vspace*{1mm}\\
$^1$CSE, Michigan State University, USA, ~
$^2$CS, Wayne State University, USA,\\
$^3$ECE, University of Minnesota, USA,~
$^4$MIT-IBM Watson AI Lab, USA \vspace*{1mm}

}

\maketitle
\vspace*{-5em}

\begin{abstract}
Recently, bi-level optimization (BLO)  has taken center stage in some very exciting developments in the area of signal processing (SP) and machine learning (ML). Roughly speaking, BLO is a classical optimization problem that involves two levels of hierarchy (\textit{i.e.}, upper and lower levels), wherein obtaining the solution to the upper-level problem requires solving the lower-level one. BLO has become popular largely because it is powerful in modeling problems in SP and ML, among others, that involve optimizing {\it nested} objective functions. Prominent applications of BLO range from resource allocation for wireless systems to adversarial machine learning. In this work, we focus on a class of {\it tractable} BLO problems that often appear in SP and ML applications. We provide an overview of some basic concepts of this class of BLO problems, such as their optimality conditions, standard algorithms (including their optimization principles and practical implementations), as well as how they can be leveraged to obtain state-of-the-art results for a number of key SP and ML applications. Further, we discuss some recent advances in BLO theory, its implications for applications, and point out some limitations of the state-of-the-art that require significant future research efforts. Overall, we hope that this article can serve to accelerate the adoption of BLO as a generic tool to model, analyze, and innovate on a wide array of emerging SP and ML applications. 
\end{abstract}


\IEEEpeerreviewmaketitle

\section{Introduction}
\label{sec: Intro}

Bi-level optimization (\textbf{BLO}) is a class of optimization problems involving two nested levels ({\em upper} and {\em lower-level}),  where  the objective and variables
of the {\it upper-level} problem depend on the optimizer of the {\it lower-level} one. The canonical formulation of {BLO} is given by:

\vspace*{-5mm}
{\small
\begin{align}
    \overbrace{  \displaystyle \minimize_{\boldsymbol \theta \in \mathcal U}    ~~  f(\boldsymbol \theta, \boldsymbol \phi^*(\boldsymbol \theta)), }^\text{Upper-level optimization over $\btheta$} ~~~~
       \st  ~~ \underbrace{ \displaystyle \boldsymbol \phi^*(\boldsymbol \theta)\in \argmin_{ h(\boldsymbol{\theta},\boldsymbol \phi)\le 0 } g(\boldsymbol \theta, \boldsymbol \phi) }_\text{Lower-level optimization over $\boldsymbol \phi$},
     \tag{{BLO}}
     \label{eq: prob_BLO}
\end{align}}%
where we assume that 
$f$, $g$ and $h$ are bivariate {\it smooth} functions,
$\btheta \in \mathbb R^m$ denotes the upper-level variable subject to the upper-level constraint set $\mathcal U$, $\boldsymbol \phi \in \mathbb R^n$ is the lower-level variable subject to the constraint $h(\boldsymbol{\theta},\boldsymbol \phi)\le 0$ that couples both $\boldsymbol{\theta}$ and $\bphi$, and $\boldsymbol \phi^*(\boldsymbol \theta)$ is one lower-level optimal solution. It is evident that the lower-level problem is an {\it auxiliary} problem since its solution supports the upper-level problem in finding a better solution.

The study of BLO  can be traced to that of Stackelberg games \cite{von1952theory}, where the upper (resp. lower) problem optimizes the action taken by a leader (resp. the follower). Early works in optimization utilized BLO to solve resource allocation problems \cite{bracken1973mathematical,  bracken1974defense, bracken1978production}; see \cite{colson2007overview} for a comprehensive survey of BLO algorithms up until mid-2000, and some more recent surveys on discrete BLO \cite{sinha2017review}, BLO under uncertainty \cite{beck2022survey}, and nonlinear and nonconvex aspects of BLO \cite{beck2022computationally}. In recent years, BLO has regained popularity because a subclass of BLO has been used to formulate and solve various challenging problems in signal processing (\textbf{SP}), machine learning (\textbf{ML}), and artificial intelligence (\textbf{AI}). 
Notable applications in SP include resource management \cite{sun2018learning,chen2020stackelberg,gao2020stackelberg,sun2022learning},  signal demodulation  \cite{simeone2018very, park2020learning, park2022predicting}, channel prediction \cite{abdi2002space, simeone2004lower, cicerone2006channel}, image reconstruction \cite{crockett2022bilevel} and image denoising 
\cite{crockett2021motivating}. In addition, BLO has also been used to make ML models, especially deep neural networks (\textbf{DNNs}), robust \cite{zuo2021adversarial,zhang2022revisiting,yang2021robust,zugner2018adversarial,robey2023adversarial,huang2020metapoison,munoz2017towards}, generalizable \cite{finn2017model,rajeswaran2019meta,andrychowicz2016learning,chen2022learning,ravi2016optimization, nichol2018first,behl2019alpha,arjovsky2019invariant,ahuja2020invariant,rosenfeld2020risks,lin2022bayesian, zhou2022sparse}, efficient
\cite{borsos2020coresets,zhao2021dataset,wang2018dataset,cazenavette2022dataset,zhang2022advancing}, easier to train
\cite{liu2020generic,franceschi2018bilevel,franceschi2017forward} and scalable \cite{liu2018darts,xu2019pc,jiang2020sp,wong2018transfer,elsken2019neural,liu2018progressive,xue2021rethinking}.

As can be easily imagined, the popularity of BLO in aforementioned applications is largely attributed to its ability to handle (often implicit)  \textit{hierarchical structures}. 
To better illustrate the challenges brought by the hierarchical architecture, an example application of \textit{coreset selection}  for model training \cite{raghu2021meta,borsos2020coresets, sun2022learning} is discussed below.

\begin{examplebox}{\textbf{Motivating application: Coreset selection for model training}}
Many contemporary SP and ML applications are facing significant challenges in data storage, transportation, and computation because they have to deal with excessive amounts of data.
Consequently, the task of identifying the most informative subset of data from a larger pool becomes crucial \cite{raghu2021meta, borsos2020coresets, sun2022learning}. This leads to the problem of {\it coreset selection}, which consists of two tasks: (T1) selecting the most representative data samples to form the coreset, and (T2) validating the performance of the selected coreset in model training. More specifically, the problem can be formulated as follows:

\vspace*{-5mm}
{\small
\begin{align}
\displaystyle \minimize_{\mathbf w \in \mathcal{U}}         & \,\,  
\ell_{\mathrm{val}}({\btheta^\ast}(\mathbf w)) \tag{T1} \label{eq: dataset_prune_UL} \\
\st & \,\, \btheta^\ast(\mathbf w) = \displaystyle \argmin_{\btheta} \ell_{\mathrm{tr}}(\btheta, \mathbf w), \tag{T2}
\label{eq: dataset_prune}
\end{align}
}%
where $\mathbf{w}$ represents the weight vector for data selection, with $w_i = 0$ indicating that the $i^\text{th}$ data sample is not selected. These weights are subject to the sparsity constraint $\mathcal{U}$, such as $\|\mathbf{w}\|_1 \leq k$, with $k$ being the selection budget. The model parameters trained on the selected data points are denoted by $\btheta$. The training loss for the model $\btheta$ with the data selection scheme $\mathbf w$ is denoted by $\ell_{\mathrm{tr}}$, while the validation loss measuring the performance of the learned model $\btheta^\ast(\mathbf w)$ over the coreset is denoted by $\ell_{\mathrm{val}}$. The above BLO formulation is also related to the data reweighting problem \cite{holtz2022learning} and hyperparameter optimization \cite{shaban2019truncated}.
\end{examplebox}

Clearly, the coreset selection  is a typical BLO problem, where the upper-level and lower-level tasks are intertwined: without knowing the training result, it is hard to gauge how representative the selected dataset is effective, while without having the coreset, one cannot perform model training to evaluate the utility of the learned model further.  More importantly, these two tasks form a {\it hierarchy}, with the model training problem being the main optimization problem, while the data selection problem is an {\it auxiliary} problem that supports the training.

Given the growing interest in BLO, this work presents an overview of a class of \textit{tractable}  \eqref{eq: prob_BLO} problems that hold significant importance in SP and ML. Roughly speaking, the class of BLO problems we consider have some desirable properties (to be discussed shortly) that allow the development of efficient and practical algorithms. We will discuss the basic concepts for this class of BLO problems along with their optimality conditions, standard algorithms (including their theoretical properties and practical implementations), as well as how they can be used to obtain state-of-the-art results for a number of key SP and ML applications.

In the existing literature, several recent surveys have been conducted on general BLO problems \cite{dempe2020bilevel,beck2022survey, beck2022computationally}. However, these surveys primarily focus on the mathematical foundations of BLO through a classical optimization lens. 
Other works \cite{ukkusuri2013bi, chen2022gradient, liu2021investigating} aim to provide comprehensive reviews of BLO algorithms, but they lack an in-depth discussion on the proper recognition and utilization of BLO, specifically addressing `when' and `how' to apply it effectively. The most relevant work to ours is \cite{liu2021investigating}, which examines complex learning and vision problems from the BLO perspective. However, the theoretical component of BLO is missing, and it overlooks a significant portion of emerging SP and ML applications (\textit{e.g.}, those discussed in Sec.\,\ref{sec: resource_allocation}-\ref{sec: general_AI}).

Different from the existing surveys on BLO 
\cite{beck2022survey, beck2022computationally,ukkusuri2013bi, chen2022gradient, liu2021investigating} to 
provide a broad overview of BLO  in its most generic form, we focus on problems that are relevant to SP and ML applications and strive to tightly integrate theory with applications, especially those that have been largely left out of the existing surveys. A few \textbf{highlights} of this article are listed below.

{First}, we distill the common structures and properties of BLO that emerge across applications related to  developing robust, parsimonious, and generalizable data-driven models in SP and ML. Our goal is to provide insights about when, where, and how BLO formulations and algorithms can be best used to yield a significant performance boost, as compared with traditional, or heuristic algorithms. In this process,  we present some  recent theoretical results about BLOs and their associated algorithms to give a flavor of the current advances in the research area, while discussing their practical and scalable implementations.

{Second}, we dive deep to understand the performance of a selected subset of state-of-the-art (SOTA) BLO algorithms on a number of representative applications. Instead of relying on results reported in existing works, which may not always be comprehensive, we designed an experiment plan and implemented all benchmarking algorithms ourselves. The goal is not only to showcase the effectiveness of the BLO-based algorithms but also to analyze the pros (\textit{e.g.}, modeling flexibility and accuracy performance) and the cons (\textit{e.g.}, run-time efficiency) of different subclasses of BLO methods.

\begin{figure}[tbh]
\centering{    
\includegraphics[width=1.0\textwidth]{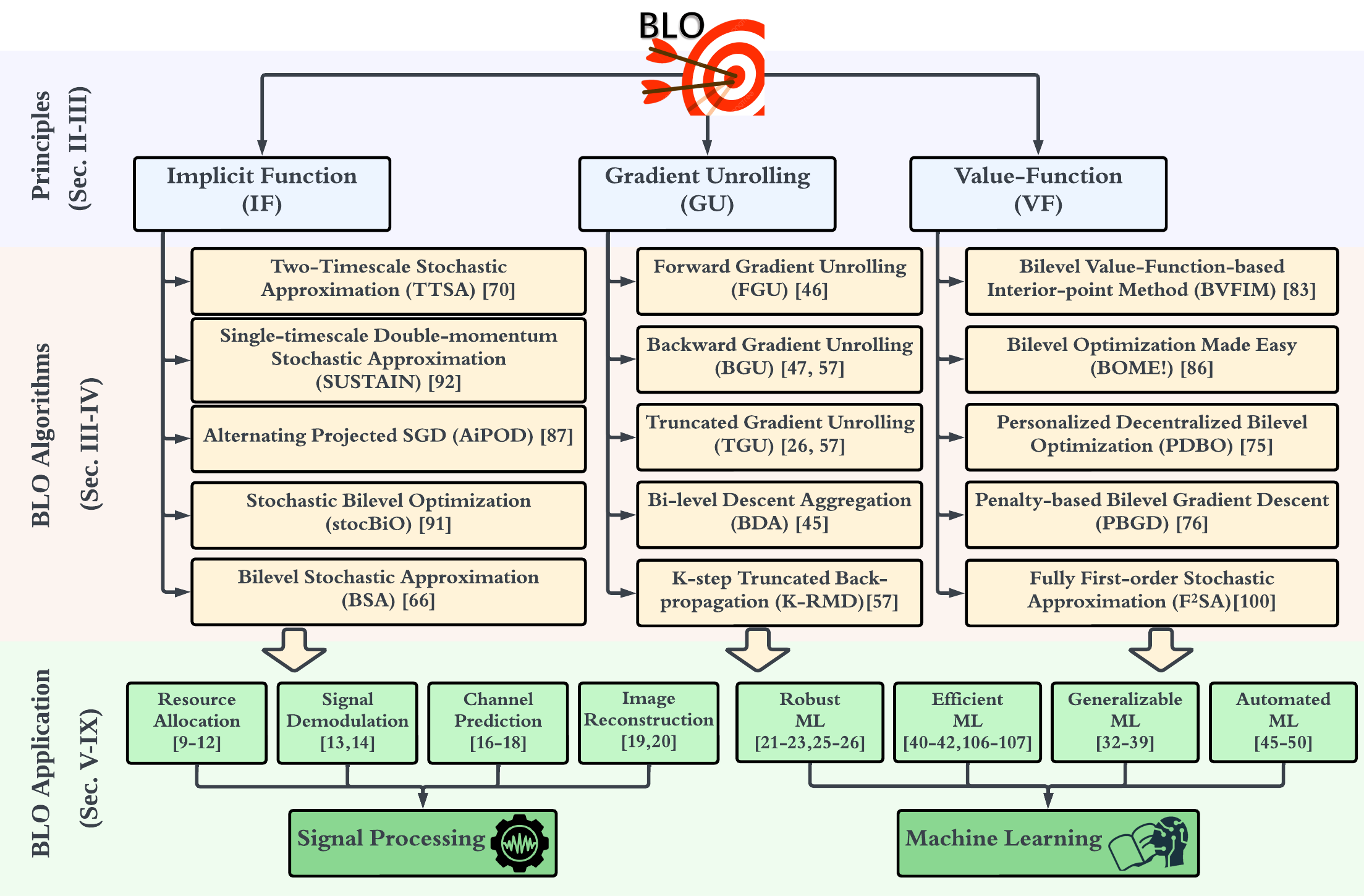}
\caption{An taxonomy of the solvers and application tasks for BLO. }
}
\label{fig: alg_app_overview}
\end{figure}

Overall, we hope that our balanced treatment of the subject will serve as the cornerstone for the accelerated adoption of BLO in diverse application areas. We hope that this tutorial can assist researchers with modeling, analyzing, and innovating on the emerging SP and ML applications. And it can serve to sparkle new theoretical and applied research to advance BLO. \textbf{Fig.\,\ref{fig: alg_app_overview}} provides an overview of the topics to be covered in this article. 

\noindent
{\em Notations:} We use lower case letters (\textit{e.g.}, $a$), lower case bold face letters (\textit{e.g.}, $\mathbf a$), and upper case bold face letters (\textit{e.g.}, $\mathbf A$) to denote scalars, vectors, and matrices, respectively. For a vector $\mathbf a$, we use $\| \mathbf a \|_p$ to denote its $\ell_p$ norm with the typical choice $p \in \{ 1,2,\infty \}$. For a matrix $\mathbf A$, we use the superscript $\top$ (or $^{-1}$) to denote the transpose (or inverse) operation.
We use $\mathbf I$ to represent the identity matrix. 
For a   function $f(\mathbf x, \mathbf y)$ (with  $\mathbf x \in \mathbb R^m$ and $\mathbf y \in \mathbb R^n$), we use $\nabla_{\mathbf x} f(\mathbf x, \mathbf y) \in \mathbb R^m$ (or $\frac{\partial  f }{\partial \mathbf x } $) and  $\nabla_{\mathbf y} f (\mathbf x, \mathbf y) \in \mathbb R^n$ (or $\frac{\partial  f }{\partial \mathbf y } $)
to denote the \textit{partial derivatives} of $f$ \textit{w.r.t.} the partial input argument $\mathbf x$ and $\mathbf y$, respectively. In contrast, we use $\frac{d f}{d \mathbf x} \in \mathbb R^m$ to represent the \textit{full derivative} of $f$ \textit{w.r.t.} $\mathbf x$, namely,
$\frac{d f}{d \mathbf x} = \nabla_{\mathbf x} f(\mathbf x , \mathbf y) + \frac{ d \mathbf y}{d \mathbf x}^\top \nabla_{\mathbf y} f(\mathbf x , \mathbf y)$ following the chain rule, where $\frac{ d \mathbf y}{d \mathbf x}^\top   \in \mathbb R^{m \times n}$ denotes the Jacobian matrix of $\mathbf y$ \textit{w.r.t.} $\mathbf x$. For ease of notation, the transpose in $\frac{ d \mathbf y}{d \mathbf x}^\top   \in \mathbb R^{m \times n}$ will be omitted if its definition is clear from the context. 
We use $\nabla_{\mathbf x} \nabla_{\mathbf y} f $ or $\nabla_{\mathbf x, \mathbf y}^2 f  \in \mathbb R^{m \times n}$ to denote the second-order partial derivative of $f$. 

\section{Warm-up: Introducing  Basic Concepts of BLO} \label{sec: basics}

\subsection{A Class of Tractable BLOs}\label{sub:tractable}

We start by discussing the challenges associated with the generic form of \eqref{eq: prob_BLO}. Even under the assumption that all involved functions are well-behaved, such as the convexity or concavity  of $f(\cdot, \cdot)$ and $g(\cdot, \cdot)$, and the linearity of  $h(\cdot,\cdot)$, solving the problem can still be highly challenging (\textit{i.e.}, {\it NP-hard}). 
To see this, let us consider the following simple example. 

\begin{examplebox}{\textbf{Example\,1: Non-convex BLO}}
  Consider the following BLO, where $f(\theta,{\phi})=-g(\theta, {\phi})=\theta^{2} - \theta \cdot {\phi} - {\phi}^{2}$:
    \begin{align*}\label{example_difficult}
   \displaystyle \minimize_{\theta \in [-1,1]} ~
    \theta^{2} - \theta \cdot {\phi}^{\ast}(\theta) - {\phi}^{\ast}(\theta)^{2}; ~~
     \st ~ {\phi}^{\ast}(\theta)= \argmin_{{\phi} \in [-1,1], \theta-{\phi} = 0} 
     -\left(\theta^{2} - \theta\cdot {\phi} - {\phi}^{2} \right).
    \end{align*}
    Notice that the objective $f(\theta,{\phi})$ (resp. $g(\theta,{\phi})$) is strongly convex (resp. strongly concave) in $\theta$, strongly concave in ${\phi}$ (resp. strongly convex), and the above BLO problem is subject to linear constraints in both the upper and lower level. In other words, both the upper and lower-level problems are convex problems. Nonetheless, it can be shown that solving the above BLO requires tackling a \textit{non-convex problem}, which in general is NP-hard. Indeed, it is not hard to see that ${\phi}^{\ast}(\theta)=\theta$. As a result, the outer function can be expressed as $\ell (\theta)=-\theta^{2}$, which is a non-convex function.
\end{examplebox}

We remark that the source of difficulty of the above problem is the {\it coupling} constraint  $ \theta - \phi =0$. If this constraint is removed, the problem will become a classical saddle-point problem, expressed below, whose global optimal solution can be easily obtained:

\vspace*{-5mm}
{\small
\begin{align}
\minimize\limits_{\theta \in [-1,1]} \maximize_{\phi \in [-1,1]} \,\,
\theta^{2} - \theta\cdot {\phi} - {\phi}^{2}.
\end{align}}%
The aforementioned examples, along with numerous others in existing survey papers like \cite{liu2021investigating, ukkusuri2013bi, chen2022gradient, dempe2020bilevel}, strongly motivate us to proceed with a {\it focused} discussion for the subset of tractable \eqref{eq: prob_BLO} problems. This subset often serves as a basis for developing practical   BLO algorithms. By focusing on this subset, we can address the specific needs of modeling SP and ML problems, which frequently demand the development of efficient, and sometimes real-time, algorithms.

To this end, we consider some special classes of \eqref{eq: prob_BLO}, with the following simplifications: 
1) The lower-level constraint set, if present, is {\it linear} that is only related to  $\boldsymbol\phi$, that is,  $h(\boldsymbol\theta, \boldsymbol\phi) = 
\mathbf A \bphi - \mathbf b$ for some matrix $\mathbf A$ and vector $\mathbf b$ of appropriate sizes; 2) The solution of the lower-level problem is a {\it singleton}, and in most cases we assume an even stronger condition that the objective function $g(\cdot, \cdot)$ is strongly convex in the second argument. With these simplifications, it is possible to show some nice properties, for example, the gradient of the upper-level objective function may exist, making algorithm design and analysis tractable. Note that there have been recent works that extend these conditions and are still able to develop efficient algorithms. We will discuss these works in Sec.\,\ref{sec: convergence}.

In summary, depending on whether the lower-level problem has constraint or not, we consider the following two classes of problems, referred to as the {\it lower-uncsontrained} (\textbf{LU}) and {\it lower-constrained} (\textbf{LC}) BLO, respectively:

\vspace*{-6mm}
{\small
\begin{align}
      \displaystyle \minimize_{\boldsymbol \theta\in\mathcal{U} }    f(\boldsymbol \theta, \boldsymbol \phi^*(\boldsymbol \theta)), ~~
       \st ~~ { \displaystyle \boldsymbol \phi^*(\boldsymbol \theta) = \argmin_{\boldsymbol \phi \in \mathbb R^{n}} g(\boldsymbol \theta, \boldsymbol \phi) }.
     \tag{{\BBLO}}
     \label{eq: prob_basic_BLO}\\
      \displaystyle \minimize_{\boldsymbol \theta\in\mathcal{U} }   f(\boldsymbol \theta, \boldsymbol \phi^*(\boldsymbol \theta)), ~~
       \st ~~ { \displaystyle \boldsymbol \phi^*(\boldsymbol \theta ) = \argmin_{\boldsymbol \phi \in \mathcal C} g(\boldsymbol \theta, \boldsymbol \phi) }.
     \tag{{\CBLO}}
     \label{eq: prob_cons_BLO}
\end{align}}%
where the set $\mathcal{C}:=\{\boldsymbol \phi \mid \mathbf{A} \boldsymbol{\phi} - \mathbf{b} \le \mathbf{0}\}$.
As will be evident in Sec.\,\ref{sec: Alg_BLO},
the presence of lower-level constrains, even in the form of linear and uncoupled constraints,
can make \eqref{eq: prob_basic_BLO} much harder to deal with than \eqref{eq: prob_cons_BLO}. 

\subsection{Connections of BLO with Game Theory}
It is important to note that BLO problems have strong ties with Stackelberg or leader-follower games \cite{von1952theory}, including Stackelberg congestion \cite{li2022differentiable} and security games \cite{kar2017trends}. These are sequential games involving two players,  the leader and the follower. The leader acts first, aiming to maximize its utility by leveraging its knowledge of the follower’s anticipated response. The follower, acting second, maximizes its utility based on the leader’s action. The connection between BLO and Stackelberg games can be bifurcated as follows: First, in certain Stackelberg games, the process of identifying a solution (\textit{i.e.}, a Stackelberg equilibrium) can be framed as a BLO. Second, BLO allows a (Stackelberg) game-theoretic interpretation, where the upper- and lower-level problems correspond to the tasks of identifying the optimal actions for the leader and the follower (\textit{i.e.}, the upper/lower-level variables), respectively. A special case of Stackelberg game is min-max optimization (\textbf{MMO}) also referred to as the saddle point problem. MMO follows a bi-level structure wherein the lower-level objective $g$ in \eqref{eq: prob_BLO}. 
is \textit{exactly opposite} of the upper-level objective function $f$ (\textit{i.e.}, $g = - f$) resulting in the following special case of BLO:

\vspace*{-5mm}
{\small
\begin{align}
    \begin{array}{ll}
    \displaystyle \minimize_{\boldsymbol \theta \in \mathcal{U} }\, \maximize_{\bphi\in\mathcal{C}}    & f(\boldsymbol \theta, \boldsymbol \phi).
     \end{array}
     \tag{MMO}
     \label{eq: prob_MMO}
\end{align}}%
In fact, MMO is much simpler to deal with than BLO, and it has been heavily studied in the SP and ML communities. We refer the interested reader to the recent tutorial \cite{razaviyayn2020nonconvex} for detailed discussion.
In addition to MMO, it is also worth noting that the algorithms reviewed in this paper are generally applicable to Stackelberg games, provided that the game's BLO formulation adheres to the assumptions of the respective algorithms.

\subsection{Implicit Gradient ({\IG})} 
As alluded previously, one important reason to consider problems \eqref{eq: prob_basic_BLO} and \eqref{eq: prob_cons_BLO} is that the objective functions of these problems are potentially {\it differentiable} \textit{w.r.t.} $\boldsymbol \theta$. Indeed, by applying the chain rule, and suppose for now that the Jacobian matrix $\frac{{d} \boldsymbol \phi^*(\boldsymbol \theta)}{ {d} \boldsymbol \theta} $ exists, we have:

\vspace*{-5mm}
{\small
\begin{align}
\frac{d f (\boldsymbol \theta, \boldsymbol \phi^*(\boldsymbol \theta))}{d \btheta }
= \nabla_{\btheta} f(\boldsymbol \theta, \boldsymbol \phi^*(\btheta))
+ {\underbrace{ \frac{{d} \boldsymbol \phi^*(\boldsymbol \theta)^\top}{ {d} \boldsymbol \theta}  }_\text{{\IG}}}
\nabla_{\boldsymbol \phi} f(\btheta, \boldsymbol \phi^*(\btheta)),
\label{eq: GD_upper}
\end{align}}%
where recall from our notation convention that 
$\nabla_{\btheta} f(\btheta , \bphi) $  and  $\nabla_{\bphi} f (\btheta, \bphi) $  
represent the \textit{partial derivatives} of $f$ \textit{w.r.t.}  the partial input arguments $\btheta$ and $\bphi$ respectively,  and  $\frac{d f}{d \btheta} $ denotes  the \textit{full derivative} of $f$ \textit{w.r.t.} $\btheta$.
For ease of notation, the transpose operation $^\top$ might be omitted in the rest of the paper.  
{We refer to the Jacobian matrix $\frac{{d} \boldsymbol \phi^*(\boldsymbol \theta)}{ {d} \boldsymbol \theta} $ as \textit{{\IG}} following \cite{gould2016differentiating}. This term is introduced to characterize the gradient of the $\argmin$-based  lower-level objective function   \textit{w.r.t} the upper-level variable $\btheta$.} 
 However, {\IG} does  {\it not} always exist for generic BLO problems. Even for \eqref{eq: prob_basic_BLO} and \eqref{eq: prob_cons_BLO}, relatively strong assumptions have to be imposed; For example, $g(\cdot, \cdot)$ needs to be strongly convex in its second argument. Further, even {\IG} exists, computing it could be quite different for the two classes of problems \eqref{eq: prob_basic_BLO} and \eqref{eq: prob_cons_BLO}. 
For the former, we will show in Sec.\,\ref{sec: Alg_BLO} that the {\IG} can be expressed in closed-form using the Implicit Function Theorem \cite{ghadimi2018approximation} based on  the first-order stationary condition of the lower-level problem, \textit{i.e.}, $\nabla_{\boldsymbol \phi} f (\boldsymbol \theta, \boldsymbol \phi^*(\btheta)) = \mathbf 0$. And this is the reason that it is referred to as ``\textit{implicit}" gradient. 
Yet, in \eqref{eq: prob_cons_BLO}, the stationary condition cannot be used  since a stationary point might violate the constraint $\boldsymbol \phi \in \mathcal C$. Therefore {\IG} generally does not admit any closed-form. Additionally, for a more restricted subset of problems \ref{eq: prob_MMO}, the influence of the {\IG}-involved term ($\mathrm{IG}\cdot \nabla_{\boldsymbol \phi} f(\btheta, \boldsymbol \phi^*(\btheta))$) in \eqref{eq: GD_upper}  can be neglected. To see this, assume that the inner problem in unconstrained, i.e., $\mathcal{C}\equiv \mathbb{R}^n$, then   $\nabla_{\boldsymbol \phi} f (\boldsymbol \theta, \boldsymbol \phi^*(\btheta)) = \mathbf 0$ based on the fact that  $\nabla_{\boldsymbol \phi} g (\boldsymbol \theta, \boldsymbol \phi^*(\btheta)) = \mathbf 0$ and $g = - f$ for MMO.

\subsection{{{BLO with Non-Singleton Lower-Level Solutions}}}
As we have mentioned in Sec. \ref{sub:tractable}, throughout this article, we will mostly focus on the case where the lower-level solution $\phi^\ast(\btheta)$ is unique, \textit{i.e.}, a singleton. Yet, if the lower-level problem involves a non-singleton (\textbf{NS}) solution, the resulting BLO problem is typically cast as

\vspace*{-8mm}
{\small
\begin{align}
    \displaystyle \minimize_{\btheta, \bphi' \in \mathcal{S}(\btheta)}  f(\btheta, \bphi') ~~
    \st ~~ \mathcal{S}(\btheta) \coloneqq \argmin_{\bphi \in \mathcal{C}} g(\btheta, \bphi)
    \tag{NS-BLO}
        \label{eq: Optimistic_BLO}
\end{align}}%
where $\mathcal{S}(\btheta)$ denotes a solution set. 
The above formulation also referred to as the {\it optimistic} BLO with non-singleton lower-level solutions, has been discussed in the literature such as \cite{sinha2017review,dempe2020bilevel}.
Note that problem \eqref{eq: Optimistic_BLO} presents significantly greater challenges from both practical and theoretical perspectives compared to problem \eqref{eq: prob_BLO}. This is because optimization over $\bphi$ is coupled across both upper- and lower-level objectives.
While our work primarily focuses on BLO with a singleton lower-level solution, we will also explore in Sec.,\ref{sec: Alg_BLO} the applicability of BLO algorithms, derived from \eqref{eq: prob_BLO}, to solve problem \eqref{eq: Optimistic_BLO}.

\begin{center}
    \rule{0.99\linewidth}{1.5pt}
\begin{center}
\vspace*{-3mm}
\textbf{\large{\textsc{Theory and Algorithms for Tractable BLOs}}}
\vspace*{-5mm}
\end{center}
\rule{0.99\linewidth}{1pt}
\end{center}

In the next two sections,
we will delve into the essential optimization principles employed in BLO, explore several popular classes of BLO algorithms, and examine their theoretical properties.

\section{Algorithmic Foundations of BLO}
\label{sec: Alg_BLO}

This section presents  an overview of three key optimization frameworks used to solve the tractable BLO problems \eqref{eq: prob_basic_BLO} and \eqref{eq: prob_cons_BLO}. The first two classes  both leverage (some approximated version of) the IG as defined in  \eqref{eq: GD_upper}. The key difference is how the approximation of the IG is conducted: One  directly assumes that there is some given procedure that can provide a high-quality solution of the lower-level problem, while the other approximates the lower-level solution by unrolling a given  algorithm for a fixed number of steps. The third class is referred to as the value function (\textbf{\VF})-based approach, which reformulates BLO  as   a single-level regularized optimization problem. It is worth mentioning that this approach offers flexibility in handling lower-level constraints  and solving non-singleton lower-level problems \eqref{eq: Optimistic_BLO}.

\subsection{The Implicit Function (\IF)-based Approach}
\label{sec: IF_alg}

\subsubsection{{\IF} for lower-level unconstrained BLO}
\label{sec: IF_UBLO}
Let us examine the problem   setup \eqref{eq: prob_basic_BLO} with a  singleton lower-level solution. For ease of theoretical analysis in Sec.\,\ref{sec: convergence},  we further assume that $g(\cdot)$ is strongly convex. In certain applications, one can explicitly add a strongly convex regularization function, such as $\gamma \times \| \boldsymbol \phi \|_2^2$ (with large enough $\gamma$), to satisfy such an assumption.

The reason that we call this approach {\it IF-based} is that we will explicitly utilize the {\it Implicit Function Theorem} \cite{krantz2002implicit} to calculate the \IG~as expressed in \eqref{eq: GD_upper}.
Recall from \eqref{eq: prob_basic_BLO} that  $\boldsymbol \phi^*(\btheta)$ is a lower-level solution, thus it satisfies:

\vspace*{-5mm}
{\small
\begin{align}
    \nabla_{\boldsymbol \phi} g (\boldsymbol \theta, \boldsymbol \phi^*(\btheta)) = \mathbf 0.
    \label{eq: stationary_UBLO}
\end{align}}%
Following the implicit function theorem, we can take the first-order derivative of \eqref{eq: stationary_UBLO} \textit{w.r.t.} $\boldsymbol \theta$, yielding
$\frac{d}{d \boldsymbol \theta} [ \nabla_{\boldsymbol \phi} g (\boldsymbol \theta, \boldsymbol \phi^*(\btheta)) ] = \mathbf 0$. Further assume that the lower-level objective $g(\cdot)$ is second-order differentiable, we can then obtain the {\IG} in \eqref{eq: GD_upper} in the following:

\vspace*{-5mm}
{\small \begin{align}
{
\frac{d \boldsymbol \phi^*(\btheta)}{d \btheta} 
}   =  - \nabla^2_{ \boldsymbol \theta, \boldsymbol \phi }g(\btheta, \boldsymbol \phi^*(\btheta)) \nabla^2_{\boldsymbol \phi, \boldsymbol \phi}g(\btheta, \boldsymbol \phi^*(\btheta))^{-1}.
\label{eq: IG_UBLO}
\end{align}}%
As observed above,  the computation of  {\IG}    involves 
the mixed (second-order) partial derivative $ \nabla^2_{ \boldsymbol \theta, \boldsymbol \phi }g$ and the inverse of the Hessian $ \nabla^2_{\boldsymbol \phi, \boldsymbol \phi} g$. Yet, computing these quantities can be challenging in practice. Therefore, the class of IF approach utilizes different kinds of approximation techniques to approximately compute the IG as expressed \eqref{eq: IG_UBLO}. We summarize the IF approach in \textbf{Algorithm\,1}.

\begin{algbox}{\textbf{Algorithm\,1: {\IF}-based approach for solving \eqref{eq: prob_basic_BLO}}}
Given initialization   $\btheta_0$,  learning rate $\alpha > 0$, and iteration number $T$;  Iteration $t\ge 0$ yields:
 
$\bullet$ \textit{Lower-level optimization:} Given $\btheta_{t}$, obtain an approximate solution of the lower-level problem, denoted as $\widetilde{\boldsymbol {\phi}}(\boldsymbol \theta_t)$; 

$\bullet$ \textit{Approximation:} Based on $\widetilde{\boldsymbol {\phi}}(\boldsymbol \theta_t)$, compute approximated versions of two second-order matrices in \eqref{eq: IG_UBLO}, denoted as $\widetilde{\nabla}^2_{ \boldsymbol \theta, \boldsymbol \phi }g(\btheta, \widetilde{\boldsymbol \phi}(\btheta_t))$ and $\widetilde{\nabla}^2_{\boldsymbol \phi, \boldsymbol \phi}g(\btheta, \widetilde{\boldsymbol \phi}(\btheta_t))^{-1}$; Compute an approximated {\IG} following \eqref{eq: GD_upper}:

\vspace*{-5mm}
{\small
\begin{align}
{\tilde{\nabla} f(\boldsymbol \theta_t)} \coloneqq \nabla_{\btheta} f(\boldsymbol \theta_t, \widetilde{\boldsymbol \phi}(\btheta_t))
- \widetilde{\nabla}^2_{ \boldsymbol \theta_t, \boldsymbol \phi }g(\btheta_t, \widetilde{\boldsymbol \phi}(\btheta_t)) \widetilde{\nabla}^2_{\boldsymbol \phi, \boldsymbol \phi}g(\btheta_t, \widetilde{\boldsymbol \phi}(\btheta_t))^{-1}
\nabla_{\boldsymbol \phi} f(\btheta_t, \widetilde{\boldsymbol \phi}(\btheta_t)).
\label{eq: GD_upper_Approx}
\end{align}}%

$\bullet$  
\textit{Upper-level optimization:} Utilize $\tilde{\nabla} f(\boldsymbol \theta_t)$ in \eqref{eq: GD_upper_Approx} to update $\boldsymbol \theta_t$ through, \textit{e.g.}, gradient descent (GD):
$
\btheta_{t+1} \leftarrow \btheta_t - \alpha \tilde{\nabla} f(\boldsymbol \theta_t)
$.
\end{algbox}

\subsubsection{Practical considerations of IF}
\label{sec: practice_IF}
In Algorithm\,1, the main computational overhead arises from the inverse-Hessian gradient product $\mathbf H^{-1} \mathbf{g}$, where $\mathbf H := \widetilde{\nabla}^2_{\boldsymbol \phi, \boldsymbol \phi}g(\btheta_t, \widetilde{\boldsymbol \phi}(\btheta_t)) $ and $\mathbf v :=  \nabla_{\boldsymbol \phi} f(\btheta_t, \widetilde{\boldsymbol \phi}(\btheta_t))$. Yet, in many contemporary applications, directly computing and storing the Hessian is computationally prohibitive. 
{To address the scalability challenge in the {\IF} method, we introduce four approaches to approximate the inverse-Hessian gradient product (or the inverse-Hessian)} $\mathbf H^{-1} \mathbf{g}$:  
the conjugate gradient (\textbf{CG}) method \cite{nazareth2009conjugate,shaban2019truncated}, the WoodFisher approximation \cite{singh2020woodfisher}, {the Neumann-series method to directly estimate the inverse-Hessian} \cite{ghadimi2018approximation,hong2020two}, and a Hessian-free simplification \cite{zhang2022revisiting,zhang2022advancing}. These methods offer different trade-offs between computational costs, with the CG method being the most computationally expensive and the Hessian-free simplification being the least expensive.

{First}, the {CG} approach  maps the product $\mathbf H^{-1} \mathbf{g}$ to the solution of a quadratic program defined as
$\min_{\mathbf x} \mathbf x^\top \mathbf H \mathbf x/2 - \mathbf g^\top \mathbf x$.  By utilizing the first-order gradient descent algorithm, we can numerically approximate
$\mathbf H^{-1} \mathbf{g}$. 
However, the convergence speed of the CG method relies on the smallest eigenvalue of the positive definite matrix $\mathbf{H}$. Therefore, if the lower-level problem is {not well-conditioned}, the CG method can be slow.
This approach has also been employed in the context of model-agnostic meta-learning (MAML)   \cite{rajeswaran2019meta} and adversarially robust training \cite{zhang2022revisiting}.

{Second}, the WoodFisher approximation \cite{singh2020woodfisher} expresses the Hessian as a recurrence of a rank-one modified Hessian estimate and calls the Woodbury matrix identity to compute the inverse of a rank-one modification to the given Hessian  matrix. 
The \textit{one-shot} WoodFisher approximation is equivalent to the quasi-Newton approximation \cite{hazan2016introduction}, $\mathbf H \approx  \mathbf v \mathbf v^T + \gamma\mathbf I$, where  $\gamma> 0$ is the dampening term  to render the invertibility of $\mathbf H$. We can then readily obtain the inverse-Hessian vector product by the Woodbury matrix identity $\mathbf H^{-1} \mathbf v = \gamma^{-1} \mathbf v + \frac{\gamma^{-2} \mathbf v \mathbf v^\top }{1 + \gamma^{-1} \mathbf v^\top \mathbf v} \mathbf v$. 
Further, an \textit{iterative} WoodFisher approximation for Hessian inverse proposed in \cite{singh2020woodfisher} enhances the estimation accuracy. 

{Third, one may utilize a Neumann-series approximation to estimate the inverse-Hessian directly by the approximation\footnote{Assuming $\mathbf H$ is normalized to ensure $\| \mathbf H\| \leq 1$.} $\mathbf H^{-1}  \approx  \sum_{i = 0}^K [\mathbf I - \mathbf H]^i$. Note that as $K \to \infty$ the approximation becomes more and more accurate \cite{horn90}. This technique is popular for approximating the inverse-Hessian in a stochastic setting wherein the upper- and lower-level objectives are accessed via a stochastic oracle \cite{ghadimi2018approximation,hong2020two}. Here, we briefly describe the procedure to approximate the inverse-Hessian stochastically using the Neumann-series method.}
{Let us choose $k$ uniformly randomly from the set $\{0,1, \ldots, K -1\}$, access batch samples of $g(\btheta, \bphi)$ denoted by $\{ g(\btheta, \bphi; \zeta_k) \}_{k =  1}^{k}$, and compute:}

{
\vspace*{-5 mm}
{\small{
\begin{align}
\label{eq: NS_Approx}
  \mathbf H^{-1} \approx \frac{k}{L_g} \prod_{i = 1}^{k} \Big( I - {\nabla^2_{\bphi, \bphi} g(\btheta, \bphi; \zeta_i) }/{L_g}  \Big)   
\end{align}}}%
where $L_g$ is the Lipschitz-smoothness constant of $g(\btheta, \bphi; \zeta_k)$.
The above procedure requires the computation of $k$ stochastic Hessians and their products. Importantly,  this estimator is a biased estimator of the inverse-Hessian with the bias decreasing exponentially with $K$ \cite[Lem.\,3.2]{ghadimi2018approximation}.}

{Finally}, to ensure the local convexity, some quadratic regularization term is usually added to the lower-level problem in BLO \cite{zhang2022revisiting,zhang2022advancing}. This modifies  \eqref{eq: prob_basic_BLO} to

\vspace*{-5mm}
{\small
\begin{align}
\displaystyle \minimize_{\boldsymbol \theta\in\mathcal{U} }    f(\boldsymbol \theta, \boldsymbol \phi^*(\boldsymbol \theta)), ~~
\st ~~ \displaystyle \boldsymbol \phi^*(\boldsymbol \theta) = \argmin_{\boldsymbol \phi \in \mathbb R^{n}} g(\boldsymbol \theta, \boldsymbol \phi) + \frac{\lambda}{2}\|\bphi\|_2^2,
\label{eq: prob_BLO_quadratic_reg}
\end{align}}%
where recall that $\lambda > 0$ is a regularization parameter. 
In this context, the Hessian-free simplification is usually adopted, which assumes $\widetilde{\nabla}^2_{\bphi, \bphi}g(\btheta_t, \widetilde{\boldsymbol \phi}(\btheta_t))  = \mathbf 0$.
This assumption can be reasonable when the lower-level objective function $g$ involves deep model training. 
For instance, in the case of a neural network with ReLU activation, the decision boundary is piece-wise linear in a tropical hyper-surface, leading to an approximate Hessian of zeros \cite{alfarra2020decision}.  This Hessian-free simplification has been used for pruning deep neural networks \cite{zhang2022advancing}. Thus, the Hessian matrix of the entire lower-level objective function in \eqref{eq: prob_BLO_quadratic_reg}  will be simplified to $\mathbf H \approx \lambda\mathbf{I}$.

\subsubsection{Extension to lower-level constrained BLO}
\label{sec: IF_CBLO}
Unlike the previous subsection, it turns out that when including constraints to the lower-level problem, the \IG~no longer has the closed-form expression because the stationary condition in \eqref{eq: stationary_UBLO} does not hold anymore. To see the impact of having constraints (even linear ones) on the lower-level problem, we present the following example, where the gradient $df/d\btheta$  is not rigorously defined. 
\begin{examplebox}{\textbf{Example\,2: \eqref{eq: prob_cons_BLO} can still  be non-differentiable} \cite{khanduri2023linearly}}
    Consider the following special case of \eqref{eq: prob_cons_BLO}, where the lower-level objective is strongly convex in both scalar variable $\theta$ and $\phi$, the upper-level is linear, and both levels are subject to linear constraints:
    \begin{align*}
        \minimize_{\theta \in [0,1]} \,\, \theta + \phi^\ast(\theta) \qquad \st\,\, \phi^\ast(\theta) \in \argmin_{1/2 \leq \phi \leq 1 } 
        (\theta - \phi)^2 
    \end{align*}
    It follows that $\phi^\ast(\theta) = 1/2$, for $\theta\leq 1/2$, and $\phi^\ast(\theta) = \theta$, for $\theta > 1/2$. 
    We notice that at the point $\theta=1/2$ the mapping $\phi^\ast(\theta)$ is continuous, but not differentiable. As a result, the outer function $\theta + \phi^\ast(\theta)$ is \textit{non-differentiable}.
\end{examplebox}

An immediate question is, can we still leverage the \IF-based approach for this subclass of problems?
It turns out that if we make some additional assumptions on the matrix $\mathbf C$ in the constraint set of \eqref{eq: prob_cons_BLO}, one can still apply the Implicit Function Theorem to the  Karush-Kuhn-Tucker (\textbf{KKT}) condition of the lower level problem to calculate the IG \cite{zhang2022revisiting,khanduri2023linearly}.
It is also important to note that {\IF}-based approaches are typically not suitable for handling general non-linear constraints in the lower-level problem. For problems with complex constraints, value function or penalty-based approaches are often employed  \cite{sow2022constrained,shen2023penalty}. 

\subsection{The Gradient Unrolling (GU)-based Approach}
\label{sec: GU_alg}
{\GU}-based approach is another class of popular algorithms for solving BLO problems in practice \cite{pearlmutter2008reverse, maclaurin2015gradient, shaban2019truncated}.  Different from the {\IF}-based framework, it employs an unrolled lower-level optimizer as an intermediary step to connect the lower-level solution with the upper-level optimization process.  The automatic differentiation (\textbf{AD}) technique is then used to compute gradients \textit{w.r.t.} the upper-level optimization variable $\btheta$. Consequently, the computation of {\IG} in {\GU} is dependent on the choice of the lower-level optimizer, and it no longer  uses the implicit function-based expression \eqref{eq: stationary_UBLO}-\eqref{eq: IG_UBLO}.

\subsubsection{{\GU}-based approach for unconstrained BLO}
\label{sec: GU_UBLO}
In particular, the {\GU}-based approach approximates $\bphi^\ast (\btheta)$ by running a given algorithm for a fixed number of iterations, and then inserting the entire trajectory into the upper-level objective; see \textbf{Algorithm\,2} for an illustration of the idea. 

\begin{algbox}{\textbf{Algorithm\,2: {\GU}-based approach for solving \eqref{eq: prob_basic_BLO}}}

Given  initialization $\bphi_0$ and $\btheta_0$, 
and iteration numbers $K$ and $T$; Let $h(\cdot): \mathcal{U}\times \mathcal{C}\to \mathcal{C}$ denote one step of a given algorithm, which 
takes both $\btheta$ and $\bphi$ as input, and outputs an updated $\bphi$. At iteration $t\ge 0$,

$\bullet$ \textit{Lower-level optimization by $K$-step optimization:}
{\small
\begin{align}
\begin{array}{ll}
     & \boldsymbol \phi_{k} = q(\btheta_t, \bphi_{k-1}), \quad k=1,\cdots K.
\end{array}
\label{eq: GU_GD}
\end{align}}%
Define  
$\widetilde{\bphi}(\btheta_t) \Def \boldsymbol \phi_K = q(\btheta_t, q(\btheta_t, \cdots, q(\btheta_t,\bphi_0)))$;

$\bullet$ \textit{Upper-level Optimization:} 
Leverage AD to compute the approximated gradient $\widetilde{\nabla} f(\btheta_t, \widetilde{\phi}(\btheta_t)) \Def 
\frac{d f(\btheta_t, q(\btheta_t, q(\btheta_t, \cdots, q(\btheta_t,\bphi_0))))}{d\btheta}$, and use this gradient to update  $\boldsymbol \theta_{t}$. 
\end{algbox}

To see the difference between the AD and the IF-based approaches, let us consider the simple case where $h(\cdot)$ is the gradient mapping, $q(\btheta_t, \bphi_{k-1}) = \boldsymbol \phi_{k-1} - \beta \times  \nabla_{\boldsymbol \phi} g (\boldsymbol \theta_t, \boldsymbol \phi_{k-1})$ (for some constant stepsize $\beta>0$), and $K=1$ (\textit{i.e.}, a single-step gradient descent step is performed for lower-level optimization). Further assume that $\boldsymbol \phi_0$ is independent of $\boldsymbol \theta$, then
the closed-form expression of {\IG} can be written as:

\vspace*{-5mm}
{\small
\begin{align}
    \displaystyle\frac{d \widetilde{\boldsymbol \phi}(\boldsymbol \theta_t)}{d \boldsymbol \theta}   = \displaystyle
    \frac{d [\boldsymbol \phi_{0} - \beta \times \nabla_{\boldsymbol \phi} g (\boldsymbol \theta_t, \boldsymbol \phi_{0}) ]}{d \boldsymbol \theta}  = -\beta \nabla_{\boldsymbol \theta, \bphi}^2 g(\boldsymbol \theta_t, \boldsymbol \phi_0).
\label{eq: IG_GU_1step}
\end{align}}%
In some sense, the above computation is simpler than the computation of $\IG$~ \eqref{eq: IG_UBLO} in the IF-based approach,  since the Hessian inverse is no longer needed. However, things can get much more complicated very quickly, as the total number of inner iterations $K$ increases. Suppose that $K=2$, we obtain:

\vspace*{-5mm}
{\small
\begin{align}
    \displaystyle\frac{d \widetilde{\boldsymbol \phi}(\boldsymbol \theta_t)}{d \boldsymbol \theta} =
    \displaystyle\frac{d [\boldsymbol \phi_1 - \beta \nabla_{\boldsymbol \phi} g (\boldsymbol \theta_t, \boldsymbol \phi_{1}) ]}{d \boldsymbol \theta}
    = -\beta\left[ \mathbf I + \beta \times \nabla^2_{\bphi, \bphi} g (\btheta_t, \bphi_1) \right ] \nabla_{\boldsymbol \theta, \bphi}^2 g(\boldsymbol \theta_t, \boldsymbol \phi_0).
\label{eq: IG_GU_2step}
\end{align}}%
Clearly, the Hessian inverse is still not needed, but as the number of unrolling steps increases, much higher computational and memory requirements will be involved.

\subsubsection{Practical considerations}
\label{sec: practice_GU}

When the unrolling step $K$ becomes too large or the problem scale itself is computationally expensive for {\GU}, manual unrolling becomes necessary to save memory costs and reduce computational overhead. Various GU approaches have been proposed to achieve this goal efficiently. Notable techniques include forward gradient unrolling (\textbf{FGU}) \cite{franceschi2017forward}, backward gradient unrolling (\textbf{BGU}) \cite{maclaurin2015gradient, shaban2019truncated, franceschi2017forward, franceschi2018bilevel}, and truncated gradient unrolling (\textbf{TGU}) \cite{luketina2016scalable, shaban2019truncated, huang2020metapoison}.
In FGU, unrolling is performed iteratively, and in the final step $K$, the Jacobi of $\phi_K$ w.r.t. $\theta$ is given by:

\vspace*{-5mm}
{\small
\begin{align}
    \underbrace{\frac{d \bphi_{K}}{d \btheta}}_{\BZ_K} = \underbrace{\frac{\partial \bphi_K}{\partial \bphi_{K-1}}}_{\BA_{K}}  \underbrace{\frac{d \bphi_{K-1}}{d \btheta}}_{\BZ_{K-1}} + \underbrace{\frac{\partial \bphi_K}{\partial \btheta}}_{\BB_K} = \cdots
    = \sum_{i=1}^{K} \left (  \prod_{t=i+1}^{K} \BA_t \right) \BB_i + \left( \prod_{t=1}^{K} \cdot \BA_t \right) \BZ_0 
    \label{eq: FGU_derivation}.
\end{align}}%
We also assume that $\bphi_0$ is independent of $\btheta$, which implies $\frac{d \bphi_0}{d \btheta}=0$.
Consequently, the expression \eqref{eq: FGU_derivation} can be rewritten as the following iterative form:

\vspace*{-5mm}
{\small
\begin{align}
        \BZ_k = \BA_k \BZ_{k-1} + \BB_k, ~k = 1, 2, \cdots, K.
        \tag{FGU}
        \label{eq: FGU}
\end{align}}%
{Both $\BA_k$ and $\BB_k$ will be calculated along with the $k$-th lower-level step $\bphi_k = q(\btheta_t, \bphi_{k-1})$ and will be discarded immediately after $\BZ_k$ is obtained. Such an iterative} nature of \eqref{eq: FGU} makes it particularly suitable for scenarios that involve a large number of unrolling steps $K$, {as the memory cost of calculating $\BA_k$ and $\BB_k$ only involves any gradient flow generated within the $k$th step.}
However, \eqref{eq: FGU} requires keeping track of the matrices $\BA_k$, $\BB_k$, and $\BZ_{k-1}$. Hence, it may not be suitable for problems with high-dimensional variables $\btheta$ and $\bphi$.

To achieve more efficient computations when $\btheta$ and $\bphi$ are of large scale, BGU is introduced, which eliminates the need for storing any intermediate matrices \cite{maclaurin2015gradient, shaban2019truncated, franceschi2017forward, franceschi2018bilevel}. BGU explores the calculation of the {\IG} following  \eqref{eq: GD_upper}:

\vspace*{-5mm}
{\small
\begin{align}
& \frac{d f(\btheta,\bphi_K)}{d\btheta} = 
\underbrace{\frac{\partial f(\btheta, \bphi_K)}{\partial \btheta} }_{\bc_K} + 
        {\underbrace{\frac{d \bphi_K^\top}{d \btheta}}_{\BZ_K^\top}}  
        \underbrace{\frac{\partial f(\btheta, \bphi_K)}{\partial \bphi_K}}_{\bd_K} 
        \overset{\eqref{eq: FGU}}{=} \bc_K + (\BZ_{K-1}^\top\BA_K^\top + \BB_K^\top) \bd_K \nonumber \\ 
        = & \underbrace{( \bc_K + \BB_K^\top\bd_K)}_{\bc_{K-1}} + \BZ_{K-1}^\top \cdot \underbrace{\BA_{K}^\top\bd_K}_{\bd_{K-1}} 
        = \bc_{K-1} +  \BZ_{K-1}^\top \bd_{K-1} 
        = \cdots 
        = \bc_{0} + \BZ_{0}^\top \bd_{0} = \bc_{-1}.
        \label{eq: bgu_general}
    \end{align}
}%
{Instead of calculating {\IG} explicitly like \eqref{eq: FGU} does, \eqref{eq: bgu_general} directly obtains the gradient of the upper-level variable, which can be further simplified the following recursive formulas:}

\vspace*{-5mm}
{\small
\begin{align}
\begin{array}{l}
\bc_{k-1}  = \bc_{k} + \BB_k^\top \bd_k, ~k= 0, 1, \cdots, K, \text{ with } \bc_K = \displaystyle\frac{\partial f(\btheta, \bphi_K)}{\partial \btheta} \\ 
\bd_{k-1} = \BA_t^\top \bd_k, ~ k= 0, 1, \cdots, K,  \text{ with } \bd_K = \displaystyle\frac{\partial f(\btheta, \bphi_K)}{\partial \bphi_K}. 
\end{array}
\tag{BGU}
\label{eq: BGU}
\end{align}
}%
It can be observed that \eqref{eq: BGU} only requires storing vectors {($\bc_k$ and $\bd_k$)} throughout the recursion by utilizing the Jacobian-vector product trick. As a result, BGU is particularly advantageous for problems with large-scale variables compared to \eqref{eq: FGU}. {Yet, due to the recursive nature, BGU can be conducted only after all the $K$ lower-level steps are finished.} Thus, \eqref{eq: BGU} needs to store all the unrolling steps $\{\bphi_k \in \mathbb{R}^n\}_{k=1}^K$, compared with {FGU}.
Consequently, it may not be efficient to handle BLO as the number of unrolling steps grows $K$.
 
It should also be noted that {\GU}   differs from {\IF} as its computation relies on the choice of the lower-level optimizer. 
For instance, employing sign-based gradient descent (signGD) \cite{bernstein2018signsgd, liu2018signsgd} as the lower-level optimizer leads to a computationally efficient GU variant referred to as signGD-based GU \cite{fan2021sign}. Specifically, the modified lower-level update rule \eqref{eq: GU_GD} becomes:

\vspace*{-5mm}
{\small
\begin{align}
\widetilde{\bphi}(\btheta_t) = \boldsymbol \phi_K; ~~ \boldsymbol \phi_{k} = \boldsymbol \phi_{k-1} - \beta \mathrm{sign} \left (  \nabla_{\boldsymbol \phi} g (\boldsymbol \theta_t, \boldsymbol \phi_{k-1}) \right ),~ k = 1,2,\ldots, K,
    \tag{signGD}
    \label{eq: GU_signGD}
\end{align}}%
where $\mathrm{sign}(\cdot)$ denotes the element-wise sign operation, and $\beta > 0$ is a certain learning rate. Given the approximation $\frac{d \,\mathrm{sign( \mathbf x)}}{ d \mathbf  x} = \mathbf 0$ (holding almost everywhere), {\IG} can be simplified to:

\vspace*{-5mm}
{\small
\begin{align}
    \frac{d \widetilde{\bphi}(\btheta_t) }{d \btheta} =   \frac{d \bphi_K}{d \btheta}  = \frac{d \bphi_{K-1}}{d \btheta}  = \ldots = \frac{d \bphi_{0}}{d \btheta}.
    \label{eq: signGD_IG}
\end{align}}%
In the case that $\bphi_{0}$ is independent of $\btheta$, we can achieve the {\IG}-free variant of the {\GU} approach.

\subsection{Value Function (VF)-based Approach}
\label{sec: VF_Based} VF-based methods \cite{liu2021value, mehra2019penalty, liu2021valuefunction,shen2023penalty,  sow2022constrained,liu2022bome} can also avoid the computation of the inverse of the Hessian required in {\IF}.
The key technique is to reformulate a standard BLO problem into a constrained single-level optimization problem. This reformulation involves transforming the lower-level problem into an upper-level inequality constraint.   
The  resulting VF-based variants can then be solved using algorithms for constrained optimization.
Furthermore, in comparison to {\IF} and {\GU}-based methods, the VF-based approach has broader applicability in solving complex BLO problems. Not only can it handle lower-level objectives with non-singleton solutions (including both convex and non-convex objectives), but more importantly,  it can accommodate lower-level constraints as well.
However, the VF-based approach has not been popular in practical SP and ML applications yet, mainly because this approach has not been able to  deal with large-scale {\it stochastic} problems. This point will be illustrated shortly in Sec. \ref{sec: convergence}.

To understand VF-based approach, consider the following {\it equivalent} reformulation of \eqref{eq: prob_cons_BLO}:

\vspace*{-6mm}
{\small
\begin{align}
   \displaystyle \minimize_{\btheta, \bphi \in \mathcal{C}} f(\btheta, \bphi), \quad  \st ~ g(\btheta, \bphi) \leq g^\ast(\btheta),
    \label{eq: vf_ori_vf}
\end{align}}%
where $g^\ast(\btheta) \coloneqq \min_{\bphi \in \mathcal{C}} g(\btheta, \bphi)$ is referred to as the  VF (\textbf{value function})  of the lower-level problem. 
{However, solving \eqref{eq: vf_ori_vf} is highly non-trivial, partly because  $g^\ast(\btheta)$ is not necessarily smooth, and can be non-convex.
To address these challenges, a relaxed version of problem \eqref{eq: vf_ori_vf} is typically considered by replacing $g^\ast(\btheta)$ with a smooth surrogate \cite{liu2021value, sow2022constrained}}:

\vspace*{-5mm}
{\small
\begin{align}
\label{Eq: VF_Reformulation}
    g^\ast_\mu(\btheta) & =  \displaystyle \minimize_{\bphi \in \mathcal{C}} g(\btheta, \bphi) + \frac{\mu_1}{2} \|\bphi\|_2^2 + \mu_2,
\end{align}}%
where $\mu \coloneqq (\mu_1, \mu_2)$ is a pair of positive coefficients that are introduced to guarantee the smoothness of $g^\ast_\mu(\btheta)$  and to ensure the feasibility of the inequality constraint $g(\btheta, \bphi) \leq g^\ast_\mu(\btheta)$.

Given the relaxed VF formulation, one can adopt standard non-linear optimization algorithms, such as a penalty-based algorithm to solve the constrained optimization problem \eqref{eq: vf_ori_vf}. For example, a log-barrier interior-point method named BVFIM is leveraged in \cite{liu2021value} to solve a sequence of smooth approximated single-level problems of \eqref{eq: vf_ori_vf}. Other related methods to solve \eqref{eq: vf_ori_vf} include PDBO \cite{sow2022constrained}, BOME \cite{liu2022bome}, and V-PBGD \cite{shen2023penalty}.

\section{Theoretical Results of BLO}
\label{sec: convergence}

In this section, we examine the theoretical guarantees of various BLO optimization methods. The section is divided into two parts. In the first part, we discuss the convergence  results of popular algorithms  for solving the \eqref{eq: prob_basic_BLO} problem, while the second part will focus on the more general formulations such as \eqref{eq: prob_cons_BLO} and \eqref{eq: Optimistic_BLO}. We list specific algorithms that can handle both stochastic and deterministic BLO problems. Given the differences in theoretical analysis between stochastic and deterministic optimization, we also consider a generalized stochastic version of BLO, whose   upper-  and lower-level objectives are

\vspace*{-5mm}
{\small
\begin{align}
\begin{array}{ll}
 \displaystyle  f( \btheta,  \bphi^*(\btheta)) \coloneqq \frac{1}{N} \sum_{i = 1}^{N} f( \btheta, \bphi^*(\btheta) ; \xi_i) 
 , \quad 
  & 
  \displaystyle g( \btheta,  \bphi) \coloneqq \frac{1}{N} \sum_{i = 1}^{N}g(\btheta,  \bphi; \zeta_i) 
,
     \end{array}
\label{eq: prob_BLO_finitesample}
\end{align}}%
where   $\xi_i \sim \mathcal{D}_f$ (resp. $\zeta_i \sim \mathcal{D}_g$)  represents the data sample of the upper-level (resp. lower-level) objective from the distribution $\mathcal{D}_f$ (resp. $\mathcal{D}_g$), and $N$ is the total number of data samples. 

\subsection{Convergence Measures of BLO}
\label{sec: convergence_measure}

In what follows, 
we introduce the convergence measures utilized for evaluating the performance of BLO algorithms. These measures serve to assess the quality of solutions obtained by these algorithms.
{First, we focus on the concept of an \textbf{$\epsilon$-stationary} point for \eqref{eq: prob_basic_BLO}, which plays a crucial role in characterizing the convergence properties of BLO when the upper-level problem is unconstrained, \textit{i.e.}, $\mathcal{U} = \mathbb{R}^m$.}
In the deterministic setting of \eqref{eq: prob_basic_BLO}, a point $\bar{\btheta} \in \mathbb{R}^m$ is considered an {$\epsilon$-stationary point} if it satisfies $ \| \nabla  f(\bar{\btheta},\bphi^\ast(\bar{\btheta}))    \|_2^2 \leq \epsilon$. 
In the stochastic setting \eqref{eq: prob_BLO_finitesample}, where the algorithm incorporates randomness, the expectation is taken over the stochasticity of the algorithm. Thus, an $\epsilon$-stationary point is defined as $\mathbb{E} \|  \nabla f(\bar{\btheta},\bphi^\ast(\bar {\btheta}))  \|_2^2 \leq \epsilon$.
{It is important to note that when the upper-level problem in \eqref{eq: prob_basic_BLO} is constrained, \textit{i.e.}, $\mathcal U \subset \mathbb{R}^m$
then the upper-level objective   $f(\bar{\btheta},\bphi^\ast(\bar{\btheta}))$ may not be differentiable over $\bar{\btheta}$ in general.  When solving \eqref{eq: prob_cons_BLO} using the {\IF}-based approach, if the {\IF} is differentiable, similar measures of stationarity as in standard optimization can be utilized. However, if the {\IF} is non-differentiable, alternative stationarity measures are commonly employed. These include subgradient optimality \cite{khanduri2023linearly}, proximal gradient methods \cite{xiao2022alternating, shen2023penalty}, and Moreau envelope techniques \cite{khanduri2023linearly,hong2020two,chen2022singletimescale}.
Furthermore, when employing VF-based approaches to solve \eqref{eq: prob_cons_BLO}, a widely used measure of stationarity is to evaluate the convergence of the algorithms towards the KKT points of the constrained reformulation of the BLO problem \cite{xiao2023generalized, sow2022constrained}.
}

In addition, the concept of \textbf{oracle complexity} is employed to quantify the number of gradient evaluations needed to obtain an $\epsilon$-stationary solution, as defined earlier. We denote $\mathcal{G}(f, \epsilon)$ (resp. $\mathcal{G}(g, \epsilon)$) as the total number of (stochastic) gradients of $f$ (resp. $g$) evaluated to achieve an $\epsilon$-stationary solution. This measure provides insights into the computational requirements of BLO algorithms and their scalability with respect to the problem size and desired solution accuracy.

\subsection{Convergence Guarantees for \ref{eq: prob_basic_BLO}}

As discussed in Sec.\,\ref{sec: Alg_BLO}, solving \eqref{eq: prob_basic_BLO} requires computing the Hessian or its inverse. However, the stochastic formulation \eqref{eq: prob_BLO_finitesample} leads to some challenges, especially for convergence analysis.
For example, in the case of solving the stochastic BLO problem  using Algorithm\,1 ({\IF}), the gradient estimates would be replaced with appropriate stochastic gradient estimates for both the upper and lower-level updates. However, obtaining an {\it unbiased} estimator for  the Hessian inverse term in {\IG}  (as defined in \eqref{eq: IG_UBLO}) is challenging. {To overcome this challenge, a {\it biased} stochastic gradient estimator based on Neumann-series approximation as discussed in Sec.\,\ref{sec: practice_IF} has been used in \cite{ghadimi2018approximation, hong2020two}. We note that the bias of the estimator can be easily controlled by choosing a larger batch to compute the Hessian of the lower-level objective \cite[Lem.\,3.2]{ghadimi2018approximation}. Moreover, we point out that stochastic CG (discussed in Sec.\,\ref{sec: practice_IF}) can also be utilized to obtain the inverse-Hessian gradient vector product to approximate the IG \cite{arbel2022amortized, ji2021bilevel}.}

In addition, a key design choice for BLO algorithms is whether the inner problem is solved {\it accurately} or not. In a \textbf{single-loop} algorithm, one only performs a {\it fixed} number of steps for the lower-level updates before every upper-level updates \cite{hong2020two,khanduri2021near,chen2021tighter}, while in a \textbf{double-loop}, many lower-level updates are carried out to obtain a very accurate approximation of $\bphi^\ast(\btheta)$ \cite{ghadimi2018approximation,ji2021bilevel,arbel2022amortized}. Typically, the former is simpler to implement in practice, while the latter is easier to analyze since the error caused by approximating $\bphi^\ast(\btheta)$ can be well managed. {In addition, the stochastic descent direction to solve both (or either) of upper and lower-level problems can be constructed using either vanilla stochastic gradient descent ({\bf SGD}) \cite{ghadimi2013stochastic} or variance reduced ({\bf VR}) algorithms \cite{cutkosky2019momentum,fang2018spider}. Specifically, it is well known that VR-based algorithms can lead to improved theoretical convergence of stochastic algorithms compared to vanilla SGD-based algorithms to solve standard optimization problems. The VR-based algorithms accomplish this improved convergence by computing additional stochastic gradients on optimization variables computed in consecutive iterations \cite{cutkosky2019momentum,fang2018spider}. Similar behavior is observed in solving BLO algorithms using VR-based gradient constructions, as discussed next.}

\begin{algbox}{\textbf{Algorithm\,3: (S)GD and VR for solving stochastic \eqref{eq: prob_basic_BLO} and \eqref{eq: prob_cons_BLO}}}
Given initialization   $\btheta_0$ and iteration number $T$;   Iteration $t\ge 0$ yields:
 
$\bullet$ \textit{Lower-level optimization:} Given $\btheta_{t}$, call SGD (or VR based on both $\btheta_{t}$ and $\btheta_{t-1}$) to obtain a lower-level solution $\widetilde{\boldsymbol {\phi}}(\boldsymbol \theta_t)$; 
 
$\bullet$ \textit{Approximation:} Given $\widetilde{\boldsymbol {\phi}}(\boldsymbol \theta_t)$, compute a stochastic gradient estimate of \eqref{eq: GD_upper_Approx} as follows: 
\begin{itemize}
\item[--] Estimate stochastic versions of  $\nabla_{\btheta} f( \btheta_t, \widetilde{\boldsymbol \phi}(\btheta_t))$, \!$\nabla_{\boldsymbol \phi} f(\btheta_t, \!\widetilde{\boldsymbol \phi}(\btheta_t))$,  $\widetilde{\nabla}^2_{  \btheta, \boldsymbol \phi }g(\btheta_t, \!\widetilde{\boldsymbol \phi}(\btheta_t))$;
\item[--] Approximate Hessian inverse, $ \widetilde{\nabla}^2_{\boldsymbol \phi, \boldsymbol \phi}g(\btheta_t, \widetilde{\boldsymbol \phi}(\btheta_t))^{-1}$, {following \eqref{eq: NS_Approx} in Sec.\,\ref{sec: practice_IF}}
\item[--] Obtain stochastic estimate of \eqref{eq: GD_upper_Approx} and construct a descent direction ${\tilde{\nabla} f(\boldsymbol \theta_t; \tilde{\xi})}$ for $\btheta$
\end{itemize}
$\bullet$  
\textit{Upper-level optimization:} Call SGD (or VR) to update $\btheta_t$ using ${\tilde{\nabla} f(\boldsymbol \theta_t; \tilde{\xi})}$ (or ${\tilde{\nabla} f(\boldsymbol \theta_t; \tilde{\xi})}$ and ${\tilde{\nabla} f(\boldsymbol \theta_{t-1}; \tilde{\xi})}$).

\end{algbox}

{In Algorithm\,3, we provide a generic stochastic algorithm to solve BLO problems using the IF-based approach. As pointed out earlier, for double-loop algorithms $\widetilde{\bphi}(\btheta_t)$ will approximate ${\bphi^\ast}(\btheta_t)$ closely, while for a single-loop algorithm $\widetilde{\bphi}(\btheta_t)$ could be given by a crude approximation of ${\bphi^\ast}(\btheta_t)$.} \textbf{Table\,\ref{tab: stochastic_BLO}} provides a summary of the oracle complexities of existing BLO algorithms for achieving an $\epsilon$-stationary point in solving problem \eqref{eq: 
 prob_basic_BLO}. 
{The theoretical results are categorized based on  {three algorithmic families}: stochastic BLO \eqref{eq: 
 prob_BLO_finitesample}, VR-based BLO, and deterministic BLO.}
The convergence performance is evaluated in terms of the oracle complexities $\mathcal{G}(f, \epsilon)$ and $\mathcal{G}(g, \epsilon)$, as introduced in Sec.\,\ref{sec: convergence_measure}. In {Table\,\ref{tab: stochastic_BLO}}, we also illustrate the BLO solver employed by different optimization principles (\textit{i.e.}, IF, GU, or VF) and the algorithmic design choices, such as double versus single-loop.

\begin{table}[t]
\begin{center}
\caption{{Summary of convergence results of representative BLO algorithms for solving \eqref{eq: prob_basic_BLO}. The convergence performance is measured by the oracle complexity $\mathcal{G}(f, \epsilon)$ and $\mathcal{G}(g, \epsilon)$ for upper- and lower-level optimization, respectively. 
`Double' or `Single' refers to the choice of the algorithm design, either double loop or single loop.}}
\label{tab: stochastic_BLO}
\resizebox{1.0\textwidth}{!}{
\begin{tabular}{c|c|c|cc|c|c|c|cc|c|c|c|cc}
\toprule[1pt]
\midrule
\multicolumn{5}{c|}{{\bf Stochastic BLO}}
& \multicolumn{5}{c|}{{\bf Variance Reduced BLO}}
& \multicolumn{5}{c}{{\bf Deterministic BLO}} \\ 
\midrule
\multirow{2}{*}{\bf Algorithm} &\multirow{2}{*}{\bf  Principle}
 & \multirow{2}{*}{\bf  Design} 
 & \multicolumn{2}{c|}{\bf Oracle Complexity}
 & \multirow{2}{*}{\bf Algorithm} 
 &\multirow{2}{*}{\bf  Principle}
 & \multirow{2}{*}{\bf  Design} 
 & \multicolumn{2}{c|}{\bf Oracle Complexity}
 & \multirow{2}{*}{\bf Algorithm} 
 &\multirow{2}{*}{\bf  Principle}
 & \multirow{2}{*}{\bf  Design}  
 & \multicolumn{2}{c}{\bf Oracle Complexity}
\\
&
&
& {$\mathcal{G}(f, \epsilon)$}
& {$\mathcal{G}(g, \epsilon)$}
&
&
&
& {$\mathcal{G}(f, \epsilon)$}
& {$\mathcal{G}(g, \epsilon)$}
& 
&
&
& {$\mathcal{G}(f, \epsilon)$}
& {$\mathcal{G}(g, \epsilon)$}
\\ \midrule

{\textsc{BSA}} \cite{ghadimi2018approximation} 
& IF
& Double
& $\mathcal{O}(1/\epsilon^2)$ 
& $\mathcal{O}(1/\epsilon^3)$

& {\textsc{STABLE}} \cite{chen2022singletimescale}
& IF
& Single
& $\mathcal{O}(1/\epsilon^2)$ 
& $\mathcal{O}(1/\epsilon^2)$ 

& {\textsc{BA}} \cite{ghadimi2018approximation}
& IF
& Double 
& $\mathcal{O}(1/\epsilon)$ 
& $\mathcal{O}(1/\epsilon^{5/4})$  

\\

{\textsc{TTSA}} \cite{hong2020two}
& IF
& Single
& $\mathcal{O}(1/\epsilon^{5/2})$
& $\mathcal{O}(1/\epsilon^{5/2})$

& {\textsc{SUSTAIN}} \cite{khanduri2021near}
& IF
& Single
& $\mathcal{O}(1/\epsilon^{3/2})$
& $\mathcal{O}(1/\epsilon^{3/2})$

& {\textsc{AID-BiO}} \cite{ji2021bilevel}
& IF
& Single  
& $\mathcal{O}(1/\epsilon)$
& $\mathcal{O}(1/\epsilon)$ 

\\
{\textsc{stocBiO}} \cite{ji2021bilevel} 
& IF
& Double
& $\mathcal{O}(1/\epsilon^2)$
& $\mathcal{O}(1/\epsilon^2)$

& {\textsc{VRBO} \cite{yang2021provably}}
& IF
& {Double}
& {$\mathcal{O}(1/\epsilon^{3/2})$}
& {$\mathcal{O}(1/\epsilon^{3/2})$}

& {\textsc{ITD-BiO}} \cite{ji2021bilevel}
& IF
& Double  
& $\mathcal{O}(1/\epsilon)$
& $\mathcal{O}(1/\epsilon)$

\\

{\textsc{ALSET}} \cite{chen2021tighter} 
& IF
& Single
& $\mathcal{O}(1/\epsilon^2)$
& $\mathcal{O}(1/\epsilon^2)$

& {\textsc{SABA}} \cite{dagreou2022framework}
& IF
& Double 
& $\mathcal{O}(N^{2/3}/\epsilon)$
& $\mathcal{O}(N^{2/3}/\epsilon)$

& {\textsc{MSTSA}} \cite{khanduri2021momentumassisted}
& IF
& Single 
& $\mathcal{O}(1/\epsilon)$
& $\mathcal{O}(1/\epsilon)$ 

\\
{\textsc{F$^2$SA}} \cite{kwon2023fully} 
& VF
& Single
& $\mathcal{O}(1/\epsilon^{7/2})$
& $\mathcal{O}(1/\epsilon^{7/2})$

& {\textsc{F$^3$SA}} \cite{kwon2023fully} 
& VF
& Single
& $\mathcal{O}(1/\epsilon^{5/2})$
& $\mathcal{O}(1/\epsilon^{5/2})$

& {\textsc{K-RMD}} \cite{shaban2019truncated}
& GU
& Double 
& $\mathcal{O}(1/\epsilon^2)$
& $\mathcal{O}(K/\epsilon^2)$ 

\\
{\textsc{AmIGO}} \cite{arbel2022amortized}
& IF
& Double
& $\mathcal{O}(1/\epsilon^2)$
& $\mathcal{O}(1/\epsilon^2)$

& {\textsc{SBFW}} \cite{akhtar2021projection}
& IF
& Single
& $\mathcal{O}(1/\epsilon^4)$
& $\mathcal{O}(1/\epsilon^2)$

& {\textsc{FGU/BGU}} \cite{franceschi2018bilevel}
& GU
& Double 
& N/A 
& N/A  
\\
\midrule
\bottomrule[1pt]
\end{tabular}}
\end{center}
\end{table}

\paragraph{Stochastic BLO}
{This set of algorithms updates the lower- and upper-level variables using  SGD, following the {\IF} or {\GU} optimization principle. The stochastic gradient estimates are evaluated as discussed in Algorithm\,3 while for the deterministic setting, the gradients are approximated using the techniques discussed in Sec.\,\ref{sec: Alg_BLO}.} Bi-level Stochastic Approximation (\textbf{BSA}) \cite{ghadimi2018approximation} was the first algorithm that offered finite-time convergence guarantees for solving unconstrained stochastic BLO problems. It employed a double-loop algorithm where the lower-level variable is iteratively estimated with multiple SGD updates, resulting in a larger oracle complexity of $\mathcal{O}(1/\epsilon^3)$ for the inner-level optimization compared to $\mathcal{O}(1/\epsilon^2)$ for the upper-level optimization.
Two-Timescale Stochastic Approximation (\textbf{TTSA}) \cite{hong2020two}, a fully single-loop algorithm (with  projected SGD update for upper-level constrained optimization) improved the lower-level oracle complexity to $\mathcal{O}(1/\epsilon^{5/2})$, however, at the cost of worsening the upper-level oracle complexity to $\mathcal{O}(1/\epsilon^{5/2})$. 
More recently, Stochastic Bi-level Optimizer (\textbf{stocBiO}) \cite{ji2021bilevel}, Stochastic Bi-level Algorithm (\textbf{SOBA}) \cite{dagreou2022framework}, and Alternating Stochastic Gradient Descent (\textbf{ALSET}) \cite{chen2021tighter} algorithms are developed to achieve $\mathcal{O}(1/\epsilon^2)$ complexity for both the upper- and lower-level optimization. Note that stocBiO \cite{ji2021bilevel} requires a batch size of $\mathcal{O}(1/\epsilon)$ to achieve this complexity while ALSET \cite{chen2021tighter} and SOBA \cite{dagreou2022framework} rely on only $\mathcal{O}(1)$ batches to achieve the same performance. In \cite{kwon2023fully}, the authors developed Fully First-order Stochastic Approximation (\textbf{F$^2$SA}), a VF-based algorithm to solve \eqref{eq: prob_basic_BLO}. The algorithm achieved an oracle complexity of $\mathcal{O}(1/\epsilon^{7/2})$ for both upper and lower-level objectives while circumventing the need to compute Hessians (or Hessian vector products) during the execution of the algorithm. 

\paragraph{Variance-reduced (VR) stochastic BLO}
{Several VR algorithms have been proposed to improve the performance of vanilla stochastic algorithms by computing additional stochastic gradients in each iteration (see Algorithm\,3).} Examples including Single-Timescale Stochastic Bilevel Optimization (\textbf{STABLE}) \cite{chen2022singletimescale} and Momentum-Assisted Single-Timescale Stochastic Approximation (\textbf{MSTSA}) \cite{khanduri2021momentumassisted} 
utilized momentum-based variance reduction techniques \cite{cutkosky2019momentum,tran2019hybrid} for upper-level optimization, improving the performance of TTSA \cite{hong2020two} and BSA \cite{ghadimi2018approximation} and achieving a oracle complexity of $\mathcal{O}(1/\epsilon^2)$ for both upper- and lower-level objectives. 
Single-timescale Double-momentum Stochastic Approximation (\textbf{SUSTAIN}) \cite{khanduri2021near}, Momentum-Based Recursive Bilevel Optimizer (\textbf{MRBO}) \cite{yang2021provably}, and Stochastic Variance-Reduced Bilevel Method (\textbf{SVRB}) \cite{guo2021randomized} further improved the performance by applying variance reduction to both upper- and lower-level optimization, achieving the oracle complexity of $\mathcal{O}(1/\epsilon^{3/2})$.
Variance-Reduced Bilevel Optimizer (\textbf{VRBO}) \cite{yang2021provably} employed  Stochastic Path-Integrated Differential Estimator (SPIDER) \cite{fang2018spider}, a double-loop variance reduced gradient estimator, to achieve the same complexity. 
More recently, Stochastic Average Bilevel Algorithm (\textbf{SABA}) \cite{dagreou2022framework} was developed, applying SAGA (an incremental gradient estimator) \cite{defazio2014saga} achieved an oracle complexity of $\mathcal{O}(N^{2/3}/\epsilon)$, where $N$ is the number of empirical data points. In addition to F$^2$SA, the authors of \cite{kwon2023fully} also proposed Faster Fully First-order Stochastic Approximation (\textbf{F$^3$SA}) that improved on the oracle complexity of F$^2$SA.  F$^3$SA utilized momentum-based variance reduction to achieve an oracle complexity of $\mathcal{O}(1/\epsilon^{5/2})$ for both upper and lower-level objectives. 

\paragraph{Deterministic BLO}
Popular approaches for solving deterministic BLO problems include Iterative Differentiation (\textbf{ITD})  and Approximate Implicit Differentiation (\textbf{AID}) methods. 
ITD-based approaches, proposed in \cite{franceschi2017forward,franceschi2018bilevel}, established asymptotic convergence guarantees and were extended to TGU (truncated gradient unrolling) in \cite{shaban2019truncated}, achieving oracle complexities of $\mathcal{O}(1/\epsilon^2)$. 
Improved guarantees were later shown in \cite{ji2021bilevel}, achieving oracle complexities of $\mathcal{O}(1/\epsilon)$, comparable to solving a deterministic single-level optimization problem. 
AID-based approaches for solving deterministic BLO problems include Bi-level Approximation (\textbf{BA}) \cite{ghadimi2018approximation}, Approximate Implicit Differentiation Bi-level optimizer (\textbf{AID-BiO}) \cite{ji2021bilevel}, and MSTSA \cite{khanduri2021momentumassisted}. BA, a double-loop algorithm, was the first to establish convergence guarantees for deterministic BLO using AID. BA achieved an oracle complexity of $\mathcal{O}(1/\epsilon)$ for the upper-level optimization and $\mathcal{O}(1/\epsilon^{5/4})$ for the lower-level optimization. The performance of BA was improved in AID-BiO \cite{ji2021bilevel} and MSTSA \cite{khanduri2021momentumassisted}, which achieved an oracle complexity of $\mathcal{O}(1/\epsilon)$ for both upper- and lower-level objectives. Please refer to Table \ref{tab: stochastic_BLO} for a summary of the discussed approaches.

\subsection{Convergence Guarantees for  \eqref{eq: prob_cons_BLO} and \eqref{eq: Optimistic_BLO}}
\label{sec: theory_BLO_complex}
Obtaining convergence guarantees for BLO algorithms become more challenging when solving more complex problems, such as those involving lower-level constraints in \eqref{eq: prob_cons_BLO} or non-singleton lower-level solutions in \eqref{eq: Optimistic_BLO}.{
{In the previous section, the majority of the algorithms discussed employed an IF-based approach to solve the BLO problem. However, in this section,  only the algorithms designed to solve \eqref{eq: prob_cons_BLO}   utilize IF-based approaches.
It is also  worth mentioning that IF-based approaches are not applicable for solving \eqref{eq: Optimistic_BLO} (or \eqref{eq: prob_cons_BLO} with general constraints) due to the inapplicability of the implicit function theorem in this context.
Instead, standard approaches to solve these more complex problems} include interior-point methods \cite{liu2021value}, primal-dual methods \cite{sow2022constrained}, dynamic barrier gradient descent \cite{liu2022bome}, and penalty based gradient descent \cite{shen2023penalty}. As pointed out earlier, a major drawback of these algorithms is that they are exclusively developed for deterministic problems and lack efficient implementations for stochastic formulations \eqref{eq: prob_BLO_finitesample}. Consequently, they are not well-suited for large-scale SP and ML applications which often involve learning over large volumes of data.}

In the following, we present a summary of recent theoretical advancements for solving highly complex BLO problems such as \eqref{eq: prob_cons_BLO} and \eqref{eq: Optimistic_BLO} in \textbf{Table\,\ref{tab: constrained_BLO}}.
{We list the oracle complexities of representative methods, design principles, and the type of constraints present in the lower-level objective function. Note that in   Table\,\ref{tab: constrained_BLO}, we list the oracle complexity for only upper-level objective with the notion of stationarity defined according to either squared norm of the projected gradient \cite{shen2023penalty, xiao2023generalized,zhang2022revisiting,khanduri2023linearly} or KKT conditions \cite{liu2022bome,sow2022constrained}.}

\subsubsection{Theoretical results for \eqref{eq: prob_cons_BLO}}
{BLO problems of the form \eqref{eq: prob_cons_BLO} involving linear constraints of the form $\mathcal{C}\coloneqq\{\boldsymbol \phi \mid h(\btheta, \bphi) \le \mathbf{0}\}$ with $h(\btheta, \bphi) \coloneqq \mathbf{A} \bphi - \mathbf{b}$, in the lower-level have gained popularity in both theory and practice \cite{zhang2022revisiting,xiao2022alternating, khanduri2023linearly}. Under some regularity assumptions on upper- and lower-level objectives and the constraint set of the lower-level problem, IF-based methods can be developed for solving such problems. These algorithms follow the same structure as the one presented in Algorithm\,1,  with the key difference that the construction of the (stochastic) gradient estimate depends on the lower-level constraints.}

\begin{algbox}{ \textbf{Algorithm 4: SIGD, an {\IF}-based approach for \eqref{eq: prob_cons_BLO}
}}

Given initialization   $\btheta_0$ and iteration number $T$;   Iteration $t\ge 0$ yields:

$\bullet$ Call Algorithm\,1 and use the following procedure to compute the \IG

$\bullet$ {\bf Notations:} $\overline{\mathbf{A}}$ be the matrix that contains the rows of $\mathbf{A}$ that correspond to the  active constraints of inequality $\mathbf{A} \bphi  - \mathbf{b} \leq \mathbf{0}$, $\overline{\boldsymbol{\lambda}}^{\ast}(\btheta)$ is the Lagrange multipliers vector that corresponds to the active constraints at ${\bphi}^{\ast} (\btheta)$ 

\vspace*{-5mm}
{\small
\begin{align}\label{eq: IG_LC}
   \text{IG}: ~ \frac{d \bphi^{\ast}(\btheta)^\top}{d \btheta}  = &\big[ \nabla_{\bphi \bphi}^2 g(\btheta,\bphi^{\ast}(\btheta)) \big]^{-1}    \cdot \big[-\nabla_{\btheta \bphi}^2 g(\btheta,\bphi^{\ast}(\btheta)) -\overline{\mathbf{A}}^{T} \nabla \overline{\boldsymbol{\lambda}}^{\ast}(\btheta)\big] 
\end{align}}%
\vspace{-8mm}
{\small
\begin{align*} 
     \nabla \overline{\boldsymbol{\lambda}}^{\ast}(\btheta) =  -  & \big[ \overline{\mathbf{A}}\big[\nabla_{\bphi \bphi}^{2}g(\btheta,\bphi^{\ast}(\btheta)) \big]^{-1}\overline{\mathbf{A}}^{\top} \big]^{-1} \big[\overline{\mathbf{A}}\big[\nabla_{\bphi \bphi}^2 g(\btheta,\bphi^{\ast}(\btheta))\big]^{-1} \nabla_{\btheta \bphi}^2 g(\btheta,\bphi^{\ast}(\btheta))\big].
\end{align*}}%
\end{algbox}

{
{In \cite{khanduri2023linearly}, the Smoothed Implicit Gradient (\textbf{SIGD}) approach was developed to handle \eqref{eq: prob_cons_BLO} with linear inequality constraints in the lower level. SIGD is an implicit gradient descent algorithm that ensures the differentiability of the implicit function through perturbation-based smoothing. The SIGD algorithm and the expression for {\IG} used in \cite{khanduri2023linearly} is stated in \eqref{eq: IG_LC}.}
{SIGD guarantees asymptotic convergence to a stationary point. A similar IF-based approach was also utilized in \cite{zhang2022revisiting}, termed Fast Bi-level Adversarial Training (Fast-BAT). Fast-BAT aims to enhance the robustness of deep learning models against adversarial attacks using BLO. And it achieves an oracle complexity of $\mathcal{O}(1/\epsilon)$ under certain smoothness assumptions.}
} In \cite{xiao2022alternating}, the stochastic BLO problem with linear equality constraints in both the upper- and lower-level problems is considered. The authors proposed an {\IF}-based approach by constructing an approximate stochastic implicit gradient for linearly constrained BLO. They also proposed the Alternating Implicit Projected SGD (\textbf{AiPOD}) algorithm, an alternating projection method that achieves an oracle complexity of $\mathcal{O}(1/\epsilon^2)$ for both upper- and lower-level objectives. The work \cite{sow2022constrained} considered BLO with general constraints in both upper- and lower-level objectives and non-singleton lower-level solutions. {It utilized the {VF}-based approach developed in Bi-level Value-Function-based Interior-point Method (\textbf{BVFIM}) \cite{liu2021value} for solving \eqref{eq: Optimistic_BLO}} and proposed a Primal-Dual Bi-level Optimizer (\textbf{PDBO}), a primal-dual algorithm for solving \eqref{eq: vf_ori_vf} when the value function is approximated using \eqref{Eq: VF_Reformulation}. Under the assumptions of convex and compact constraint sets and convex lower-level objectives, PDBO achieves an oracle complexity of $\mathcal{O}(1/\epsilon^{3/2})$. Recently, the authors in \cite{shen2023penalty} proposed Penalty-Based Bilevel Gradient Descent (\textbf{PBGD}) for BLO with general constraints and non-singleton lower-level solutions. The authors established the equivalence of BLO and its penalty-based reformulations based on VF and KKT conditions. PBGD achieved an oracle complexity of $\mathcal{O}(1/\epsilon^{3/2})$ for solving constrained BLO with lower-level objectives satisfying the PL inequality. Note that the algorithms PDBO \cite{sow2022constrained} and PBGD \cite{shen2023penalty} can be utilized to solve both \eqref{eq: prob_cons_BLO} and
\eqref{eq: Optimistic_BLO} problems. {Next, we discuss specific algorithms for solving \eqref{eq: Optimistic_BLO}.}
 
\begin{table}[t]
\begin{center}
\caption{{Convergence results of representative  algorithms  for solving \eqref{eq: prob_cons_BLO} and \eqref{eq: Optimistic_BLO} problems. 
Similar to Table\,\ref{tab: stochastic_BLO}, convergence is measured by oracle complexity. Other algorithmic details, including optimization principles ({\IF}, {\GU}, and {\VF}), problem setups (\ref{eq: prob_cons_BLO} and \ref{eq: Optimistic_BLO}),     objective function types, and lower-level constraint types.}} 
\label{tab: constrained_BLO}
\resizebox{.8\textwidth}{!}{
\begin{tabular}{c| c | c c|c |c}
\toprule[1pt]
\midrule
\multirow{2}{*}{\bf Algorithm} 
& 
\multirow{2}{*}{\bf Principle}
&
\multicolumn{2}{c|}{\bf Objective functions}
& 
\multirow{2}{*}{\bf  Constraints} 
&
\multirow{2}{*}{\bf \begin{tabular}[c]{@{}c@{}}\textbf{Oracle} \\ \textbf{Complexity}
\end{tabular}}
 \\
& & {\bf Upper-Level} & {\bf Lower-Level} &   &  
\\
\midrule
{\textsc{BDA}} \cite{liu2020generic} 
& GU
& Strongly-Convex  & Convex
& No  & Asymptotic
 \\
{\textsc{BVFIM}} \cite{liu2021value}
& VF
& Smooth  & Non-Convex  
& No  
& Asymptotic
\\
{\textsc{BOME!}} \cite{liu2022bome} 
& VF
& Smooth  &  PL
& No  & $\mathcal{O}(1/\epsilon^{4})$\\
{\textsc{SIGD}} \cite{khanduri2023linearly}
& IF
& Smooth  & Strongly-Convex
& Linear Inequality  & Asymptotic
\\
{\textsc{AiPOD}}
\cite{xiao2022alternating}
& IF
& Smooth  & Strongly-Convex
& Linear Equality  
& $\mathcal{O}(1/\epsilon^{2})$
\\
{\textsc{PDBO}} \cite{sow2022constrained} 
& VF
& Smooth  & Convex
& Non-Linear  & $\mathcal{O}(1/\epsilon^{3/2})$
\\
\textsc{PBGD} \cite{shen2023penalty} & VF  & Smooth & PL & Non-Linear &  $\mathcal{O}(1/\epsilon^{3/2})$ \\
\midrule
\bottomrule[1pt]
\end{tabular}}
\end{center}
\end{table}

\subsubsection{Theoretical results for \eqref{eq: Optimistic_BLO}}
An attempt to relax the lower-level singleton assumption for the lower-level problem was made in \cite{liu2020generic} with the introduction of Bi-level Descent Aggregation (\textbf{BDA}), a bi-level descent framework for solving \eqref{eq: Optimistic_BLO}. BDA assumes convexity of the lower-level objective and strong convexity of the upper-level objective with respect to $\bphi$. The framework updates the lower-level variable, $\bphi$, using a convex combination of the upper- and lower-level partial gradients, and then updates the upper-level variable, $\btheta$, using standard FGU/BGU techniques. The authors established the asymptotic convergence of BDA in \cite{liu2020generic}. In \cite{liu2021value}, the authors relaxed the convexity assumptions on the lower and upper-level objectives in \eqref{eq: Optimistic_BLO} and proposed BVFIM, a VF-based approach to solving the problem. BVFIM solves a sequence of penalty-based reformulations of the VF problem using the interior-point method, with the asymptotic convergence. In \cite{liu2022bome}, the authors introduced Bi-level Optimization Made Easy! (\textbf{BOME!}), an alternative approach to directly solve the VF problem using a dynamic barrier gradient descent algorithm. Under the assumption of PL inequality for the lower-level objective, BOME! achieves a finite-time sample complexity of $\mathcal{O}(1/\epsilon^4)$ in the worst case.

\begin{center}
    \rule{0.99\linewidth}{1.5pt}
\begin{center}
\vspace*{-3mm}
\textbf{\large{\textsc{BLO-Enabled  SP and ML Applications}}}
\vspace*{-5mm}
\end{center}
\rule{0.99\linewidth}{1pt}
\end{center}

\begin{table}[htb]
\centering
\caption{An overview of emerging applications of BLO in SP and ML ({\recommend}~indicates applications we studied).
}
\label{tab: application_overview}
\resizebox{\textwidth}{!}{%
\begin{tabular}{c|c|c|c}
\toprule[1pt]
\midrule
\multirow{2}{*}{\begin{tabular}[c]{@{}c@{}}\textbf{Representative} \\ \textbf{Applications}\end{tabular}}
& \multirow{2}{*}{\begin{tabular}[c]{@{}c@{}}\textbf{Application} \\ \textbf{Areas}\end{tabular}}
& \multirow{2}{*}{\textbf{Problem Description}} & \multirow{2}{*}{\begin{tabular}[c]{@{}c@{}}\textbf{Selected} \\ \textbf{References} \end{tabular}} \\
& & & \\ \midrule
Wireless resource allocation~{\recommend} & 
\multirow{4}{*}{\begin{tabular}[c]{@{}c@{}}{SP}\end{tabular}} &
To allocate wireless resources optimally and maximize their utilities & \cite{sun2018learning,chen2020stackelberg,gao2020stackelberg, sun2022learning} \\
Signal demodulation~{\recommend} & 
&
To accurately estimate transmitted symbols from received baseband signals
& \cite{simeone2018very, park2020learning} \\
 Channel prediction &  &
 To predict the states of a communication channel by leveraging previous observations 
 & \cite{abdi2002space, simeone2004lower, cicerone2006channel} \\
Image reconstruction &  &
To recover images from their sparse measurements & \cite{crockett2021motivating, crockett2022bilevel} \\
\midrule
Adversarial training~{\recommend} & 
\multirow{2}{*}{\begin{tabular}[c]{@{}c@{}}{Robust ML}\end{tabular}}
&
To train an ML model with adversarial robustness against adversarial attacks & \cite{zuo2021adversarial,zhang2022revisiting,robey2023adversarial} \\
Poisoning attack generation  & 
&
To generate malicious data into the training set, creating vulnerabilities/backdoors in ML models.
& \cite{zhao2018data, huang2020metapoison,yang2021robust} \\
\midrule
Model pruning~{\recommend} & 
\multirow{2}{*}{\begin{tabular}[c]{@{}c@{}}{Efficient ML}\end{tabular}}
&
To find sparse sub-networks from a dense DNN w/o generalization loss & \cite{sehwag2020hydra, zhang2022advancing} \\
Dataset condensation & & To select a  subset or distill a condensed version of the training set w/o generalization loss  & \cite{wang2018dataset, borsos2020coresets, zhao2021dataset, chen2022gradient, jia2023robustness}\\
\midrule
Model-agnostic meta-learning~{\recommend} & 
\multirow{2}{*}{\begin{tabular}[c]{@{}c@{}}{Generalized ML}\end{tabular}} &
To train an ML model that can quickly adapt to new tasks using limited data & \cite{ravi2016optimization, finn2017model, nichol2018first, rajeswaran2019meta, behl2019alpha} \\

Invariant risk minimization~{\recommend}  & & To train an ML model with invariant features against distribution shift  & \cite{arjovsky2019invariant, ahuja2020invariant, rosenfeld2020risks, lin2022bayesian, zhou2022sparse}\\
\midrule
Neural architecture search &
\multirow{2}{*}{\begin{tabular}[c]{@{}c@{}}{Automated ML }\end{tabular}} &
To automatically optimize the architecture of DNNs for improved performance  & \cite{liu2018darts, elsken2019neural, liu2018progressive, xu2019pc, jiang2020sp, xue2021rethinking} \\ 
Hyper-parameter optimization & & To optimize the hyper-parameters and model selection schemes in an ML pipeline & \cite{franceschi2017forward, maclaurin2015gradient, franceschi2018bilevel, shaban2019truncated, liu2020generic}\\
\midrule
\bottomrule[1pt]
\end{tabular}%
}
\end{table}

In the following sections, we will showcase how BLO can be leveraged to obtain state-of-the-art results for
a number of key SP and ML applications, such as  wireless resource allocation (Sec.\,\ref{sec: resource_allocation}), signal demodulation (Sec.\,\ref{sec: signal_demodulation}), adversarial training for robustifying ML models (Sec.\,\ref{sec: robust_AI}), weight pruning for enhancing model efficiency (Sec.\,\ref{sec: efficient_AI}), and invariant representation learning for improving domain generalization   (Sec.\,\ref{sec: general_AI}). \textbf{Table\,\ref{tab: application_overview}} summarizes a number of emerging BLO application areas, together with some representative references. 

\section{BLO for Wireless Resource Allocation}
\label{sec: resource_allocation}

In this section, we explore the application of BLO in wireless communications, specifically in the context of wireless optimal resource allocation (power control) \cite{sun2018learning, sun2022learning}. The objective is to allocate power efficiently among multiple transmitter-receiver pairs to maximize some system-level performance. We consider a dynamic environment where wireless channel statistics change episodically, reflecting the variations in user behavior. This scenario is commonly observed in cellular networks, where user density and behavior vary throughout the day, such as more users in downtown areas during peak hours compared to night time.

To solve the resource allocation problem, we employ a neural network trained to predict the optimal power allocation for users based on channel information. However, neural networks often struggle when evaluated on data that deviate from the training distribution, which is the case in our scenario. To address this challenge and ensure the model's performance across episodes, we adopt a continual learning framework \cite{parisi2019continual}. We maintain a memory set, containing a representative subset of samples encountered so far, to facilitate adaptation to new episodes while preserving satisfactory performance on previous ones. The model is trained not only on the current batch of data but also using the current memory set. In what follows, we provide a detailed description of the problem and its formulation, following the formulation and approach presented in \cite{sun2022learning}.

\textbf{Formulation.}
Consider a dynamic wireless environment comprising $T$ episodes, where the channel state information (CSI) statistics remain stationary within each episode. Consider a supervised learning setting, where the $i$th data pair $(\mathbf{h}^{(i)}, \mathbf{p}^{(i)})$ consists of the CSI vector $\mathbf{h}^{(i)}$ (the feature vector) capturing the channel characteristics, and the corresponding optimal \textit{power allocation} $\mathbf{p}^{(i)}$ across the users (the label). We train a neural network $\pi(\btheta; \mathbf{h}^{(i)})$ on these data pairs, where $\btheta$ represents the model parameters and $\mathbf{h}^{(i)}$ serves as the network input, with the output being the  power allocation prediction.
Assume that the data arrives sequentially in multiple batches. Let $D_t$ denote the batch we receive at time $t$. Assume that there is a fixed-size memory set $M_t$ available, which stores representative historical data to be combined with $D_t$ for training, and it is updated when a new batch arrives. The performance of a power allocation scheme $\mathbf p$ (for a given CSI $\mathbf{h}$) is measured by the 
{weighted sum-rate} loss function $R(\mathbf{p};\mathbf{h})$ \cite[eq. (1)]{sun2022learning}.

At each time $t$, our problem involves two tasks. The first task is to train the neural network on a (weighted) set of training samples, aiming to find the optimal model parameter $\btheta$. The second task is to select the most representative subset from the available training data, which includes the samples in memory $M_t$ and the current data batch $D_t$ (denoted as $M_t \cup D_t$). 
This selected subset will be then used for training, as well as to form the new memory to be used in the next time $t+1$.
To achieve this, we introduce the variable $\boldsymbol{\lambda}$, which represents the weights associated with each sample. Higher weights are assigned to samples that are more representative or challenging, as determined by the system performance metric $R(\mathbf{p};\mathbf{h})$. These weighted samples are then selected to form the updated memory set, and they contribute more during the model training. The idea is that by focusing on training the model on these challenging samples, we can expect better performance on the remaining easier samples.
It is important to note that the model training process relies on the weights obtained from the second task. Consequently, the problem described above can be naturally formulated as the following BLO problem:

\vspace*{-5mm}
{\small
\begin{align}
\begin{array}{ll}
\displaystyle \minimize_{\boldsymbol{\theta}} 
& \displaystyle \sum_{i \in M_{t} \cup D_{t}} \lambda^{(i)}(\boldsymbol{\theta})  \ell(\boldsymbol{\theta};\mathbf{h}^{(i)},\mathbf{p}^{(i)}) \\ 
\st & \boldsymbol{\lambda} (\boldsymbol{\theta}) =\displaystyle \argmin_{\mathbf 1^T \boldsymbol{\lambda} = 1 } \sum_{i \in M_{t} \cup D_{t}} \lambda^{(i)} R(\pi(\boldsymbol{\theta};\mathbf{h}^{(i)}); \mathbf{h}^{(i)}),,
\end{array}
\label{blo_wireless_res}
\end{align}}%
where $\ell(\boldsymbol{\theta};\mathbf{h}^{(i)},\mathbf{p}^{(i)}) = \|\mathbf{p}^{(i)} - \pi(\boldsymbol{\theta};\mathbf{h}^{(i)}) \|_2^{2}$ is the MSE loss over the $i$th sample, and $M_{t} \cup D_{t}$ are the available training samples at time $t$. 
At the upper-level, we conduct supervised training   by assigning weights to the per-sample losses based on the weights obtained from the lower-level task,  which is similar to \eqref{eq: dataset_prune}. On the other hand, at the lower-level, we tackle the task of assigning weights to the samples based on their achieved rates, where higher weights are assigned to samples with lower rates. This is similar to \eqref{eq: dataset_prune_UL}.

\textbf{Methods.} 
Problem \eqref{blo_wireless_res} is a \textit{constrained BLO} problem with {linear equality constraints} \textit{w.r.t.} the lower-level variable $\boldsymbol{\lambda}$. 
Based on the problem structure, and the optimization principles and algorithms introduced in Sec.\,\ref{sec: Alg_BLO} and Sec.\,\ref{sec: theory_BLO_complex}, we utilize the {\IF}-based SIGD (Smoothed Implicit Gradient) method \cite{khanduri2023linearly} to solve \eqref{blo_wireless_res}.
However, it should be noted that the SIGD method assumes a strongly convex lower-level problem (see Table\,\ref{tab: constrained_BLO}). To ensure this property, we introduce a regularization term $(\gamma/2) 
\|\boldsymbol{\lambda}\|_2^2$, where $\gamma$ is the regularization parameter as described in Sec.\,\ref{sec: IF_CBLO}. As a classical baseline approach, we also consider transfer learning (TL). In TL, when a new data batch $D_{t}$ arrives, the current model trained on data up to time $t-1$ is fine-tuned using only $D_{t}$. This approach is motivated by the expectation that previous knowledge can be transferred to the new environment, enabling quick adaptation of the model. However, updating the model may result in a loss of prior knowledge, leading to performance degradation on the prior episodes.

\begin{figure}[t]
    \begin{center}
    \begin{tabular}{ccc}
    \includegraphics[width=.3\linewidth,height=!]{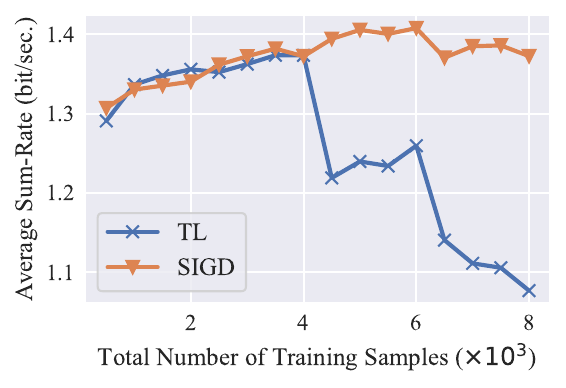} &
    \includegraphics[width=.3\linewidth,height=!]{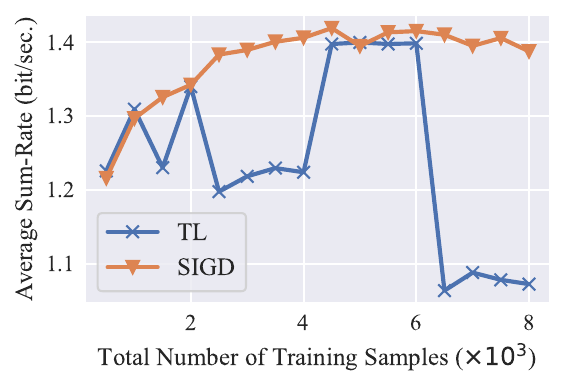} &
    \includegraphics[width=.3\linewidth,height=!]{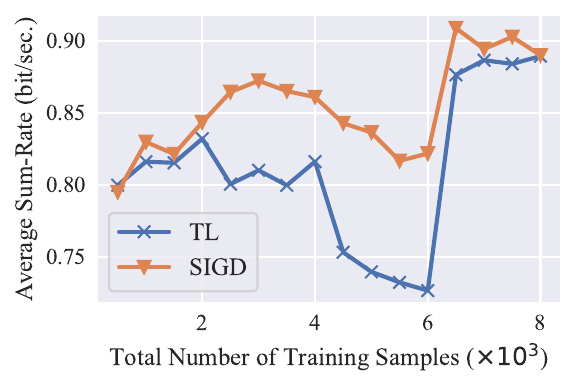}
   \vspace*{-3mm} \\
    \footnotesize{ (a)   }
    & 
   \footnotesize{ (b)  } 
   &
   \footnotesize{ (c) }
    \end{tabular}
    \end{center}
    \vspace*{-3mm}
    \caption{{
    The average sum-rate achieved on the combined test set (across all four episodes) is plotted as a function of the total no. of samples used for model training, which arrive sequentially in batches. Three experiments are conducted with different types of channel statistics across the four episodes.   (a) The channel sequence is Rayleigh-Rician-Geometry10-Geometry50.  (b) The channel sequence is Rician-Geometry10-Rayleigh-Geometry50.  (c) The channel sequence is Geometry10-Geometry20-Geometry50-Rayleigh. The numbers after the `Geometry' channel indicate the spatial arrangement of the nodes, such as a $20m \times 20m$ area for `Geometry20'.
    }}
    \label{fig: resource_allocation}
\end{figure}

\textbf {Experiment Results.} 
We consider an experiment setting with $T=4$ episodes and $10$ users, and with three different types of communication channels: Rayleigh fading, Rician fading, and Geometry channels; see \cite[pg. 15]{sun2022learning} for more details. 
The neural network trained for power allocation consists of three hidden layers with sizes $200, 80, 80$, respectively.
\textbf{Fig.\,\ref{fig: resource_allocation}} 
illustrates the performance of power allocation, measured by the sum rate; in the horizontal axis we have the total number of samples used for model training, as these arrive sequentially in batches. Here the power allocation schemes  are obtained using the BLO-based SIGD method   and  the baseline approach (TL), respectively. As we can see, the  SIGD method exhibits smoother adaptation to each episode (note that the gray line marks the boundary between episodes). It also experiences less deterioration in performance compared to the baseline approach. These results demonstrate the advantage of using BLO for power allocation.

\section{BLO for Wireless Signal Demodulation}
\label{sec: signal_demodulation}

In this section, we explore the application of BLO in wireless signal demodulation by associating it with another BLO application, model-agnostic meta-learning (\textbf{MAML}) \cite{finn2017model}. Thus, we begin by introducing the fundamental concepts of MAML within the framework of BLO, and then establish the connection between MAML and signal demodulation.

\subsection{Fundamentals of BLO in MAML}
MAML, as an optimization-based meta-learning approach, has gained significant popularity in various fields, especially in scenarios with limited resources
\cite{finn2017model,fan2021sign,balcan2019provable,khodak2019adaptive,rajeswaran2019meta,dou2019investigating,liu2020does}.
Specifically, MAML learns a meta-\textit{initialization} of {optimizee} variables  (\textit{e.g.}, model weights $\btheta$) to enable fast adaptation to new tasks when fine-tuning the model from the learned initialization with only a few new data  points~\cite{finn2017model,finn2019online}.
With $N$ learning tasks $\{ \task_i \}_{i=1}^N$, {\it (a)} a fine-tuning set $\din_i$ is used in $\task_i$  for the {\em task-specific lower-level optimization} over the task-agnostic model initialization
$\btheta$, and {\it (b)} a validation set $\dout_i$ is used in the {\em upper-level optimization} for   evaluating the fine-tuned model $\btheta_i^*$ from $\btheta$. Thus,   MAML can be formulated as the following BLO problem:

\vspace*{-5mm}
{\small
\begin{align}
\begin{array}{ll}
\displaystyle \minimize_{\btheta} & \displaystyle \frac{1}{N}\sum_{i = 1}^N \E_{(\din_i, \dout_i) \sim \task_i} \left [ \ell_{i} ( \btheta_i^* (\btheta); \dout_i)
\right ] \\
\st &
\btheta_i^*(\btheta)
= \displaystyle \argmin_{\btheta_i \in \mathbb R^n} \ell_i(\btheta_i; \din_i, \btheta),
\end{array}
\label{eq:maml-orig}
\tag{MAML-BLO}
\end{align}}%
where $\btheta_i^*(\btheta)$ signifies the fine-tuned model weights using the   initialization $\btheta$ under the task $\task_i$, and $\ell_i$ denotes the model training (or validation) loss over $\din_i$ (or $\dout_i$) with initialization $\btheta$ (or fine-tuned model $\btheta_i^*(\btheta)$).

The \ref{eq:maml-orig} problem falls into the category of unconstrained BLO. Thereby, existing works such as \cite{finn2017model, fan2021sign, Yin2020Meta-Learning,rajeswaran2019meta,fallah2020convergence} commonly  employ the {\IF} or {\GU}-based approaches to solve it. The \textit{vanilla MAML algorithm} \cite{finn2017model} utilizes GU-based BLO solver, which carries out the upper- and lower-level updates using the following steps:

\vspace*{-5mm}
{\small
\begin{align}
\hspace*{-3mm} \begin{array}{l}
\text{Lower:} ~
\btheta_i^*(\btheta) = \btheta_i^{(M)}, ~ 
\btheta_i^{(m)} \leftarrow \btheta_i^{(m-1)} - \beta \nabla_{\btheta_i}\ell_i(\btheta_i^{(m-1)}), \text{ $m = 1,\ldots, M$, given $\btheta_i^{(0)} = \btheta$}
\\
\text{Upper:} ~ \btheta \leftarrow  \btheta - \alpha \displaystyle \frac{1}{N} \sum_{i = 1}^N \nabla_{\btheta}\ell_i(\btheta_i^{*}(\btheta)),\\
\end{array}
\hspace*{-3mm}
\label{eq:maml-alg}
\tag{MAML}
\end{align}}%
where $\alpha$, $\beta>0$ represent the learning rates for the SGD updates in the upper and lower level respectively. As mentioned in Sec.\,\ref{sec: GU_UBLO}, the choice of the lower-level optimizer will greatly influence the GU-based BLO solvers.

As described in   Sec.\,\ref{sec: practice_GU}, the usage of the sign-based SGD in the lower level can lead the vanilla \ref{eq:maml-alg} to a first-order BLO solver, known as \textbf{Sign-MAML} \cite{fan2021sign}.
In contrast to GU-based MAML methods discussed above, the implicit MAML (\textbf{iMAML}) \cite{rajeswaran2019meta} utilizes IF-based approach, where the CG (conjugate gradient) method is used to compute the inverse-Hessian gradient product. Compared to the vanilla MAML, iMAML shares the same lower-level updating rule,
while adopting the {\IF}-based upper-level iteration, similar to \eqref{eq: GD_upper_Approx}.
Different from {\IF} and {\GU}, the first-order (FO) MAML (\textbf{FO-MAML}) operates by alternating between SGD-based lower-level optimization and SGD-based upper-level optimization, without explicitly considering the implicit gradient. We refer to this optimization procedure as \textbf{AO} (Alternating Optimization).
 
\begin{table}[t]
\centering
\caption{Performance comparison of various MAML methods on the commonly used datasets for few-shot learning tasks Omniglog \cite{lake2015human} and Mini-Imagenet \cite{vinyals2016matching}. The best performance in each setting is marked in \textbf{bold}. The standard deviations are reported based on 5 random trials. Rows marked in \colorbox{Gray}{gray} indicate BLO-enabled algorithms.}
\label{tab: maml_overview}
\resizebox{1.0\linewidth}{!}{%
\begin{tabular}{c|c|cc|cc|cc|cc}
\toprule[1pt]
\midrule
\multirow{2}{*}{\textbf{Method}} & \textbf{\multirow{2}{*}{\begin{tabular}[c]{@{}c@{}}\textbf{BLO}\\ (Solver)\end{tabular} }} & \textbf{Test Acc.} (\%) & \textbf{Time} (min) & \textbf{Test Acc.} (\%) & \textbf{Time} (min) & \textbf{Test Acc.} (\%) & \textbf{Time} (min) & \textbf{Test Acc.} (\%) & \textbf{Time} (min) \\ 
& & \multicolumn{2}{c|}{Omniglot 20-way-1-shot} & \multicolumn{2}{c|}{Omniglot 20-way-5-shot} & \multicolumn{2}{c|}{Mini-ImageNet 5-way-1-shot} & \multicolumn{2}{c}{Mini-ImageNet 5-way-5-shot} \\
\midrule
FO-MAML \cite{finn2017model}  & AO & 90.62\footnotesize{$\pm 0.29$} & 2.71 & 96.44\footnotesize{$\pm 0.23$} & 2.99 & 46.39\footnotesize{$\pm 0.44$} & 4.32  & 54.45\footnotesize{$\pm 0.29$} & 4.93 \\
\rowcolor{Gray}
Sign-MAML \cite{fan2021sign}  & GU & 91.75\footnotesize{$\pm 0.26$} & 2.85 & 97.79\footnotesize{$\pm 0.18$} & 2.91 & 47.73\footnotesize{$\pm 0.55$} & 4.51 & 55.12\footnotesize{$\pm 0.33$} & 4.72 \\
\rowcolor{Gray}
vanilla MAML \cite{finn2017model} & GU & 95.65\footnotesize{$\pm 0.25$} & 4.71 & 98.42\footnotesize{$\pm 0.23$} & 4.94 & 48.77\footnotesize{$\pm 0.65$} & 15.5 & \textbf{55.72}\footnotesize{$\pm 0.36$} & 15.9 \\
\rowcolor{Gray}
iMAML \cite{rajeswaran2019meta} & IF &\textbf{95.99}\footnotesize{$\pm 0.19$} & 3.64 & \textbf{98.63}\footnotesize{$\pm 0.14$} & 3.85 & \textbf{49.31}\footnotesize{$\pm 0.41$} & 11.6 & 54.71\footnotesize{$\pm 0.27$} & 13.3 \\
\midrule
\bottomrule[1pt]
\end{tabular}%
}
\end{table}

We next demonstrate the effectiveness of BLO in MAML by applying it to benchmark few-shot learning tasks on the Omniglot \cite{lake2015human} and Mini-ImageNet \cite{vinyals2016matching} datasets, where the generalization of the learned meta-initialization is evaluated on the new tasks, each with only a few examples.
We follow the standard experimental setting \cite{finn2017model,rajeswaran2019meta, raghu2019rapid, fan2021sign}, considering 20-way-1-shot and 20-way-5-shot learning on Omniglot, and 5-way-1-shot and 5-way-5-shot learning on Mini-ImageNet. Here $P$-way-$Q$-shot refers to   training a model using a small set of data points sampled from $P$ classes, with each class containing $Q$ examples.
\textbf{Tab.\,\ref{tab: maml_overview}} provides an overview of the accuracy and runtime efficiency of different methods for solving \ref{eq:maml-orig}, including AO-based first-order   MAML (FO-MAML) \cite{finn2017model}, GU-based Sign-MAML \cite{fan2021sign}, GU-based vanilla MAML \cite{finn2017model}, and IF-based iMAML \cite{rajeswaran2019meta}. As we can see,
the vanilla MAML  and iMAML generally achieve higher testing accuracy than other MAML variants, but they require more computation time.  This is expected for their fewer implementation assumptions, resulting in more precise meta initialization states. In the FO optimization category, we observe that Sign-MAML outperforms FO-MAML, demonstrating the advantage of using GU to solve BLO problems.

\subsection{BLO for Wireless Signal Demodulation}

\textbf{Background and connection to MAML.} We next examine the application of BLO in the context of wireless signal demodulation through the lens of MAML. 
Wireless signal demodulation is a critical process in the physical layer of a wireless communication system, which seeks to recover the \textit{transmitted symbols} $\mathbf s$ from the \textit{received signals} $\mathbf y$.
Our investigation aligns with prior research \cite{park2020learning} that focuses on a scenario where wireless devices frequently transmit short packets consisting of a few pilot symbols through a varying channel.
Notably, when working with any new wireless transmitter device in the field, only a limited number of pilot symbols are available to optimize the demodulator. At the same time, the historical data pairs for previous devices and channel conditions can hardly transfer to new ones. Consequently, we view this demodulation modeling as a few-shot meta-learning problem, aiming to derive a meta initialization from historical data that is able to quickly adapt to future devices.
In this context, similar to MAML,
a meta-demodulator characterized by a learnable parameter $\btheta$ is trained to quickly adjust to new devices with only a handful of new pilot symbols.

\textbf{Formulation.} Specifically, $N$ supervised datasets are collected to train the meta-demodulator, each associated with a specific device, which can be treated as $N$ tasks. The $i$-th dataset with $K$ samples is given by
$ \mathcal{D}_i = \{(\mathbf s_i^{(k)}, \mathbf y_i^{(k)})\}_{k=1}^K$,
where $\mathbf y_i^{(k)}$ represents the $k$-th received signal and $\mathbf{s}^{(k)}_i$ is its corresponding ground-truth transmitted symbol. The demodulation task for each single device can be formulated as a classification problem, as each symbol $s_i^{(k)}$ can only be one of the several binary encodings, \textit{e.g.,} ranging from $0000$ to $1111$ following the 16-Quadrature Amplitude Modulation (16-QAM) \cite{park2020learning}. Thus, the cross-entropy loss between the predicted transmitted symbol $\hat{\mathbf s}(\by, \btheta)$ and the true symbol $\mathbf s$ is used to train the demodulator model: $\ell_i(\btheta; D_i) = \E_{(\mathbf{s}_i^{(k)}, \mathbf{y}_i^{(k)}) \sim \mathcal{D}_i} \mathcal{L}_\texttt{CE}(\hat{\mathbf{s}}(\mathbf{y}_i^{(k)}, \btheta), \mathbf{s}_i^{(k)})$, where we use a multi-layer neural network as the demodulation model $\btheta$ following \cite{park2020learning}, in order to predict the transmitted symbol $\hat{\mathbf s}$ using the received signal $\mathbf{y}$.
Given $N$ devices (datasets) within the meta-training dataset, each contains $K$ data samples, which are divided into $K^{\texttt{tr}}$ for fine-tuning $\btheta$ and $K^{\texttt{val}}$ for validating the performance of the fine-tuned model. In line with the notions used in \eqref{eq:maml-orig}, the demodulation of the $i$-th device can be regarded as learning task $\mathcal{T}_i$, which consists of a fine-tuning dataset $\mathcal{D}_i^{\texttt{tr}}$ and a validation dataset $\mathcal{D}_i^{\texttt{val}}$.
To this end, we can apply the previously introduced MAML methods \cite{finn2017model,fan2021sign,rajeswaran2019meta} to address the problem of wireless signal demodulation.

\textbf{Experiment Results.} 
During the meta-training phase, we consider $N=1000$ different devices, each of which has $K^{\texttt{tr}} \in \{1, 5, 10, 20\}$ training samples designated as the fine-tuning set.
For the meta-testing phase, we use another set of $100$ devices, each of which has $K^{\texttt{tr}}$ pairs for the few-shot demodulator fine-tuning and an additional $10000$ pilot symbol pairs for symbol classification accuracy evaluation.
 \textbf{Tab.\,\ref{tab: meta_demodulate}} shows the average classification accuracy and running time for different MAML methods, providing insights into how the choice of $K^{\texttt{tr}}$ affects the performance of these methods.
As we can see, MAML and iMAML achieved the highest symbol classification accuracy, although they required more computational time, consistent with the findings in {Tab.\,\ref{tab: maml_overview}}. Sign-MAML outperformed FO-MAML in the 1-shot scenario, benefiting from the effectiveness of the {\GU} solver. However, in scenarios with more shots, FO-MAML can achieve the performance on par with MAML.

\begin{table}[t]
\centering
\caption{Performance comparison of different MAML methods for few-shot demodulation of 16-QAM modulated wireless signals. The best performance in each setting is indicated in \textbf{bold}. Standard deviations are reported based on 5 random trials.
Rows marked in \colorbox{Gray}{gray} indicate BLO-enabled algorithms.
}
\label{tab: meta_demodulate}
\resizebox{1.0\textwidth}{!}{%
\begin{tabular}{c|c|cc|cc|cc|cc}
\toprule[1pt]
\midrule
\multirow{2}{*}{\textbf{Method}} & \textbf{ \multirow{2}{*}{\begin{tabular}[c]{@{}c@{}}\textbf{BLO}\\ (Solver)\end{tabular} }} & \textbf{Test Acc.} (\%)           & \textbf{Time} (min)         & \textbf{Test Acc.} (\%)           & \textbf{Time} (min)   & \textbf{Test Acc.} (\%)           & \textbf{Time} (min)        & \textbf{Test Acc.} (\%)           & \textbf{Time} (min)        \\ 
& & \multicolumn{2}{c|}{16-way-1-shot}    & \multicolumn{2}{c|}{16-way-5-shot}  & \multicolumn{2}{c|}{16-way-10-shot}  & \multicolumn{2}{c}{16-way-20-shot}   \\
\midrule
\rowcolor{Gray}
FO-MAML \cite{finn2017model}  & AO & 94.46\footnotesize{$\pm 0.35$} & 8.59 & 97.93\footnotesize{$\pm 0.11$} & 8.73 & 99.13\footnotesize{$\pm 0.13$} & 8.87  & 99.17\footnotesize{$\pm 0.04$} & 9.01 \\
\rowcolor{Gray}
Sign-MAML \cite{fan2021sign}  & GU & 96.91\footnotesize{$\pm 0.10$} & 8.10 & 97.35\footnotesize{$\pm 0.28$} & 8.14 & 97.42\footnotesize{$\pm 0.30$} & 8.23  & 97.46\footnotesize{$\pm 0.24$} & 8.32\\
\rowcolor{Gray}
vanilla MAML \cite{finn2017model} & GU & \textbf{98.84}\footnotesize{$\pm 0.10$} & 13.04 & \textbf{99.49}\footnotesize{$\pm 0.03$} & 13.27   & \textbf{99.63}\footnotesize{$\pm 0.01$} & 13.29 & \textbf{99.66}\footnotesize{$\pm 0.02$} & 13.35\\
\rowcolor{Gray}
iMAML \cite{rajeswaran2019meta} & IF &97.66\footnotesize{$\pm 0.07$} & 14.14 & 98.58\footnotesize{$\pm 0.04$} & 14.52 & 99.35\footnotesize{$\pm 0.04$} & 14.81 & 99.43\footnotesize{$\pm 0.02$} & 14.88 \\
\midrule
\bottomrule[1pt]
\end{tabular}%
}
\end{table}

\section{BLO for Adversarially Robust Training}
\label{sec: robust_AI}

Adversarial attacks demonstrate the vulnerability of ML models to small perturbations in their inputs, posing a challenge to their reliability \cite{goodfellow2014explaining,carlini2017towards,papernot2016limitations}. As a result, developing robust models that withstand such attacks has become a critical focus in the field of robust ML. In what follows, we delve into the BLO application to improve the adversarial robustness of DNNs.

The lack of adversarial robustness in ML models has prompted extensive research on adversarial defense mechanisms \cite{madry2017towards,zhang2019theoretically,shafahi2019adversarial,wong2020fast,zhang2019you}. While most of existing defenses rely on min-max optimization (MMO) to minimize worst-case training loss by incorporating a synthesized adversary, this approach requires completely opposing objectives for the defender and attacker. This limits its applicability to scenarios where differing objectives are desired. Recent works \cite{zhang2022revisiting,robey2023adversarial} have demonstrated the use of BLO with customizable attack objectives to improve the efficiency and robustness of robust model training across a wide range of adversarial attack strengths.

\noindent \textbf{Formulation.}
We formulate the BLO-based robust training for defending against adversarial attacks. Using variables $\btheta$ for model parameters and $\bdelta$ for input perturbation, and loss functions $\ell_{\mathrm{tr}}$ for training and $\ell_{\mathrm{atk}}$ for attacks, we define the task of robust training (\textbf{RT}) as a BLO problem. The upper-level problem involves training the model $\boldsymbol{\theta}$, while the lower-level problem optimizes $\boldsymbol{\delta}$ for adversarial attack generation to produce worst-case input in model training. This yields:

\vspace*{-5mm}
{\small
\begin{align}
\begin{array}{ll}
\displaystyle \hspace*{-5mm} \minimize_{\btheta } & \hspace*{-2mm} \mathbb E_{(\mathbf x, y) \in \mathcal D} [ \ell_{\mathrm{tr}}(\boldsymbol \theta, \mathbf x +  \bdelta^*(\btheta; \mathbf x, y), y) ] \\
\hspace*{-5mm} \st  & \hspace*{-2mm} \bdelta^*(\btheta; \mathbf x, y ) = \displaystyle \argmin_{ \bdelta \in \mathcal C}  \ell_{\mathrm{atk}}(  \bdelta, \btheta; \mathbf x, y),
\end{array}
\label{eq: rob_tr_BLO}
\tag{\text{RT-BLO}}
\end{align}}%
where $(\mathbf x, y)$ is a data pair with feature $\mathbf x$ and label $y$ drawn from the training dataset $\mathcal D$, 
$(\mathbf{x} + \bdelta)$ is an adversarial example \textit{w.r.t.} $\mathbf x$, and
$\bdelta \in \mathcal C$ denotes a perturbation constraint, \textit{e.g.}, $ \mathcal{C} = \{  \bdelta \, | \, \| \bdelta \|_\infty  \leq \epsilon, \mathbf x+ \bdelta \in [\mathbf 0, \mathbf 1] \}$ for $\epsilon$-tolerated $\ell_\infty$-norm   constrained attack with normalized input in $[\mathbf 0, \mathbf 1]$.
 
If we choose $\ell_{\rm atk} = - \ell_{\rm tr}$, then problem \eqref{eq: rob_tr_BLO} reduces to the MMO-based adversarial training \cite{madry2017towards}. However, the flexibility of independently choosing the lower-level attack objective allows for a broader range of robust training scenarios. In particular, it enables the development of a fast robust training variant called Fast Bi-level Adversarial Training (\textbf{Fast-BAT}) \cite{zhang2022revisiting}.
Fast-BAT formulation, given below,  specifies the lower-level attack generation problem of \eqref{eq: rob_tr_BLO}  as a constrained  convex quadratic program:

\vspace*{-5mm}
{\small
\begin{align}
\hspace*{-5mm}  \begin{array}{ll}
\displaystyle \minimize_{\boldsymbol \theta } &
\mathbb E_{(\mathbf x, y) \in \mathcal D} [ \ell_{\mathrm{CE}}(\boldsymbol \theta,  \mathbf x + \boldsymbol \delta^*(\boldsymbol \theta; \mathbf x, y),  y) ] \\
\st & 
\boldsymbol \delta^*(\boldsymbol \theta; \mathbf x ,y) 
= \displaystyle 
\argmin_{ \boldsymbol \delta \in \mathcal C } ~ \underbrace{- (\bdelta - \boldsymbol \delta_0)^\top \nabla_{\bdelta = \boldsymbol \delta_0}\ell_{\mathrm{CE}}( \btheta, \mathbf x +  \bdelta, y)  + (\gamma/2) \| \boldsymbol \delta -\boldsymbol \delta_0 \|_2^2}_\text{$\Def \ell_{\mathrm{atk}} (\boldsymbol \theta, \boldsymbol \delta; \mathbf x, y)$}  ,
\end{array}
\hspace*{-5mm}  \label{eq: prob_biLevel_sign_lin} 
\tag{Fast-BAT}
\end{align}}%
where $\ell_{\mathrm{CE}}(\boldsymbol \theta,  \mathbf x,  y)$ is the CE (cross-entropy) loss for training model weights $\boldsymbol \theta$ evaluated at the data point $(\mathbf x, y)$, the lower-level attack objective is given by a first-order Taylor expansion   of $-\ell_{\mathrm{CE}}$ (at the linearization point $\boldsymbol \delta_0$) plus a quadratic residual with the regularization parameter $\gamma > 0$.
 Since the lower-level problem becomes a convex quadratic program,  it leads to the closed-form projected gradient descent (PGD) solution 
$\bdelta^*(\btheta) = \mathcal P_{\mathcal C} \left (
\boldsymbol \delta_0 - (1/\gamma) \nabla_{\bdelta }\ell_{\mathrm{CE}}( \btheta,  \bdelta)  \right ) \left. \right |_{\bdelta = \bdelta_0}
$.

\noindent \textbf{Methods.} As problem  \eqref{eq: prob_biLevel_sign_lin}  falls into the category of \eqref{eq: prob_cons_BLO}, we can solve it using optimization methods introduced in  Sec.\,\ref{sec: theory_BLO_complex}. 
Specifically, we consider the KKT-oriented {\IF} approach (Sec.\ref{sec: IF_CBLO}) as our BLO solver ({see Algorithm 4}). 
In our experiments, we refer to this method as \textbf{Fast-BAT-IF}.
Furthermore, we compare Fast-BAT-IF, with non-BLO representative robust training baselines, such as \textbf{Fast-AT} \cite{wong2020fast}, \textbf{Fast-AT-GA} \cite{andriushchenko2020understanding}, and \textbf{BackSmooth} \cite{chen2020efficient}.

\begin{table}[t]
\centering
\caption{{
Performance comparison of different robust training methods using PreActResNet-18 on CIFAR-10  and Tiny-ImageNet datasets. 
The training phase includes adversarial perturbations with two budgets: $\epsilon = 8/255$ and $16/255$ over 20 epochs. Results are presented as mean $\pm$ standard deviation over 10 random trials. Rows marked in \colorbox{Gray}{gray} indicate BLO-enabled algorithms.
}} 
\label{tab: adversarial_training_overview}
\begin{threeparttable}
\resizebox{0.99\textwidth}{!}{
\begin{tabular}{c|c|c|c|c|c|c|c|c}
\toprule[1pt]
\midrule
\textbf{Method} 
& \begin{tabular}[c]{@{}c@{}}
\textbf{\textbf{BLO}} \\ (\textbf{Solver})
\end{tabular}
& \begin{tabular}[c]{@{}c@{}}
\textbf{SA} (\%) \\
($\epsilon = 8/255$)
\end{tabular}
& \begin{tabular}[c]{@{}c@{}}
\textbf{RA-PGD} (\%) \\
($\epsilon = 8/255$)
\end{tabular} & 
\begin{tabular}[c]{@{}c@{}}
\textbf{RA-AA} (\%) \\
($\epsilon = 8/255$)
\end{tabular}
& \begin{tabular}[c]{@{}c@{}}
\textbf{SA} (\%) \\
($\epsilon = 16/255$)
\end{tabular} 
& 
\begin{tabular}[c]{@{}c@{}}
\textbf{RA-PGD} (\%) \\ ($\epsilon = 16/255$)
\end{tabular}
& 
\begin{tabular}[c]{@{}c@{}}
\textbf{RA-AA} (\%) \\ ($\epsilon = 16/255$)
\end{tabular}
 & \begin{tabular}[c]{@{}c@{}}
\textbf{Time} \\ (s/epoch)
\end{tabular} \\ \midrule
\multicolumn{9}{c}{{CIFAR-10, PreActResNet-18}} \\ \midrule
{{Fast-AT} \cite{wong2020fast}} 
& \multirow{3}{*}{N/A}
& \textbf{82.39}\footnotesize{$\pm 0.14$} 
& 45.49\footnotesize{$\pm 0.21$} 
& 41.87\footnotesize{$\pm 0.15$} 
& 44.15\footnotesize{$\pm 7.27$} 
& 21.83\footnotesize{$\pm 1.32$} 
& 12.49\footnotesize{$\pm 0.33$} 
& 23.1 \\

{{Fast-AT-GA} \cite{andriushchenko2020understanding}} 
&
& 79.71\footnotesize{$\pm 0.24$} 
& 47.27\footnotesize{$\pm 0.22$} 
& 43.24\footnotesize{$\pm 0.27$} 
& 58.29\footnotesize{$\pm 1.32$} 
& 26.01\footnotesize{$\pm 0.16$} 
& 17.97\footnotesize{$\pm 0.33$} 
& 75.3\\

{{BackSmooth} \cite{chen2022efficient}} 
&
& 79.31\footnotesize{$\pm 0.17$} 
& 48.06\footnotesize{$\pm 0.07$} 
& 44.55\footnotesize{$\pm 0.17$} 
& 64.88\footnotesize{$\pm 1.75$} 
& 24.18\footnotesize{$\pm 1.37$} 
& 15.47\footnotesize{$\pm 0.92$} 
& 60.6\\ \midrule
\rowcolor{Gray}
{{Fast-BAT-IF} \cite{zhang2022revisiting}}
& IF
& 79.97\footnotesize{$\pm 0.12$}
& \textbf{48.83}\footnotesize{$\pm 0.17$} 
& \textbf{45.19}\footnotesize{$\pm 0.12$} 
& \textbf{68.16}\footnotesize{$\pm 0.25$} 
& \textbf{27.69}\footnotesize{$\pm 0.16$} 
& \textbf{18.79}\footnotesize{$\pm 0.24$} 
& 61.4 \\\midrule

\multicolumn{9}{c}{{Tiny-ImageNet, PreActResNet-18}} \\ \midrule
{{Fast-AT} \cite{wong2020fast}} 
& \multirow{3}{*}{N/A}
& 41.37\footnotesize{$\pm 3.08$} 
& 17.05\footnotesize{$\pm 3.25$} 
& 12.31\footnotesize{$\pm 2.73$} 
& 31.38\footnotesize{$\pm 0.19$} 
&  5.42\footnotesize{$\pm 2.17$} 
&  3.13\footnotesize{$\pm 0.24$} 
& 284.6 \\

{{Fast-AT-GA} \cite{andriushchenko2020understanding}} 
&
& 45.52\footnotesize{$\pm 0.24$} 
& 20.39\footnotesize{$\pm 0.19$} 
& 16.25\footnotesize{$\pm 0.17$} 
& 29.17\footnotesize{$\pm 0.32$} 
&  6.79\footnotesize{$\pm 0.27$} 
&  4.27\footnotesize{$\pm 0.15$} 
& 592.7 \\

{{BackSmooth} \cite{chen2022efficient}} 
&
& 45.32\footnotesize{$\pm 0.24$} 
& 20.49\footnotesize{$\pm 0.16$} 
& 17.13\footnotesize{$\pm 0.14$}
& 31.17\footnotesize{$\pm 0.25$} 
& 7.32\footnotesize{$\pm 0.21$} 
& 4.71\footnotesize{$\pm 0.16$} 
& 511.3\\ \midrule
\rowcolor{Gray}
{{Fast-BAT-IF \cite{zhang2022revisiting}}}
& IF
& \textbf{45.80}\footnotesize{$\pm 0.22$} 
& \textbf{21.97}\footnotesize{$\pm 0.21$} 
& \textbf{17.64}\footnotesize{$\pm 0.15$} 
& \textbf{33.78}\footnotesize{$\pm 0.23$} 
& \textbf{8.83}\footnotesize{$\pm 0.22$} 
& \textbf{5.52}\footnotesize{$\pm 0.14$} 
& 572.4 \\
\midrule
\bottomrule[1pt]
\end{tabular}}
\end{threeparttable}
\end{table}

\noindent \textbf{Experiment Results.}
In \textbf{Tab.\,\ref{tab: adversarial_training_overview}}, we empirically show the performance of different robust training methods to robusify   PreActResNet-18 \cite{he2016identity}  on the CIFAR-10 \cite{Krizhevsky2009learning} and Tiny-ImageNet \cite{deng2009imagenet} datasets.
The evaluation metrics include \ding{172} the test-time {r}obust  {a}ccuracy ({RA}) of the learned model against 50-step PGD attacks \cite{madry2017towards} with 10 restarts (\textbf{RA-PGD}) using the perturbation budgets $\epsilon = 8/255$ and $16/255$;
\ding{173} RA against AutoAttack (\textbf{RA-AA})  \cite{croce2020reliable} in the setup similar to RA-PGD; \ding{174} the {s}tandard {a}ccuracy (\textbf{SA}) of the learned model on natural examples; \ding{175} the time consumption required for robust training.
As observed, Fast-BAT-IF exhibits higher robustness compared to non-BLO baselines, highlighting the effectiveness of BLO in robust training.

\section{BLO for Model Pruning}
\label{sec: efficient_AI}

While over-parameterized structures are key to the improved generalization of DNNs, they create new problems - the millions or even billions of parameters not only increase computational costs during inference but also pose serious deployment challenges on resource-limited devices. Thus, the problem of model pruning arises, aiming to  reduce the sizes of an ML model   by identifying and removing redundant model weights. 
In this section, we investigate the application of BLO in the context of model pruning
\cite{han2015deep, han2015learning,zhang2022advancing,frankle2018lottery}.

\noindent \textbf{Formulation.}
The study of model pruning through the lens of BLO was first explored in \cite{zhang2022advancing}.
Specifically, there exist two main tasks in model pruning: pruning and re-training. Pruning involves determining the sparse pattern of model weights, while re-training focuses on recovering model accuracy using the remaining non-zero weights \cite{han2015learning, frankle2018lottery}. 
To facilitate these tasks, one can introduce the binary pruning mask variable $\mathbf{m} \in \{0, 1\}^m$ and the model weight variable $\boldsymbol{\phi} \in \mathbb{R}^m$, where $m$ represents the total number of model parameters.
Accordingly, the pruned model is given by $(\mathbf{m} \odot \boldsymbol{\phi})$, where $\odot$ denotes element-wise multiplication. To achieve a pruning ratio of $p\%$, we impose a sparsity constraint on $\mathbf{m}$, where $\mathbf{m} \in \Omega$ and $\Omega = \{ \mathbf{m} ,|, \mathbf{m} \in {0, 1}^n, \mathbf{1}^T \mathbf{m} \leq k \}$, with $k = (1-p\%)n$.
Our goal is to prune the original dense model to the targeted pruning ratio of $p\%$ and obtain the optimal sparse model $(\mathbf{m} \odot \boldsymbol{\phi})$. To achieve this, we interpret the pruning task (\ding{172}) and the model retraining task (\ding{173}) as two optimization levels, leading to the formulation of bi-level pruning (\textbf{BiP}):

\vspace*{-5mm}
{\small
\begin{align}
\displaystyle \minimize_{\mathbf m \in \Omega} \underbrace{\ell_{\mathrm{tr}}(\mathbf m \odot {\boldsymbol \phi}^*(\mathbf m))}_\text{ \ding{172}: Pruning task}; ~~
\st ~~ \underbrace{\displaystyle {\boldsymbol \phi}^*(\mathbf m) = \argmin_{\boldsymbol \phi \in \mathbb{R}^n} \tilde \ell_{\mathrm{tr}}(\mathbf m\odot \boldsymbol \phi)  + \frac{\gamma}{2}  \|  \boldsymbol \phi \|_2^2}_\text{\ding{173}: Sparsity-fixed model retraining},
\label{eq: prob_pruning}
\tag{BiP}
\raisetag{5em}
\end{align}}%
where $\ell_{\mathrm{tr}}$ and $\tilde \ell_{\mathrm{tr}}$ denote the training losses  under different data batches, 
$\mathbf m$ and $\boldsymbol \phi$ are the upper-level and lower-level optimization variables respectively, and
${\boldsymbol \phi}^*(\mathbf m)$ signifies the retrained model weights given the pruning mask $\mathbf m$. 
In \eqref{eq: prob_pruning}, the lower-level training objective was regularized using a strongly convex regularizer $\frac{\gamma}{2}  \|  \boldsymbol \phi \|_2^2$ like \eqref{eq: prob_biLevel_sign_lin}.

\noindent \textbf{Methods.}
Since \ref{eq: prob_pruning} is an unconstrained BLO problem, it can be solved using     BLO algorithms, \textit{e.g.}, {\IF} and {\GU}, introduced in Secs.\,\ref{sec: IF_UBLO} and \ref{sec: GU_UBLO}. Moreover, since the Hessian of the lower-level objective function \textit{w.r.t.} model parameters  is  of  high dimension, we impose the Hessian-free assumption $\nabla_{\bphi, \bphi} \ell_{\mathrm{tr}} = \mathbf 0$ to make the BLO implementation computationally feasible. Following \eqref{eq: IG_UBLO}, one can then obtain the closed form of   {\IG} \cite{zhang2022advancing}: $\frac{d\boldsymbol \phi^*(\mathbf m)}{d\mathbf m} =  - \frac{1}{\gamma} \nabla^2_{\mathbf m, \boldsymbol \phi} \ell_{\mathrm{tr}} (\mathbf m \odot \boldsymbol \phi^*)$, where $\boldsymbol \phi^*$ signifies a lower-level solution. Furthermore,  the bi-linearity of the pruning mask $\mathbf m$ and the model weights $\boldsymbol \phi$ allows us to further simplify the {\IG} to:

\vspace*{-5mm}
{\small
\begin{align}
\displaystyle
  \frac{d\boldsymbol \phi^*(\mathbf m)}{d\mathbf m}
= -  \frac{1}{\gamma} \mathrm{diag}( \nabla_{\mathbf z} \ell_{\mathrm{tr}}(\mathbf z)
\left |_{\mathbf z = {\mathbf m} \odot {\boldsymbol \phi}^*} \right.
),
\label{eq: IG_final}
\end{align}}%
where the Hessian-free assumption is adopted and $\mathrm{diag} (\mathbf a)$ denotes the diagonal matrix with $\mathbf a$ being the main diagonal vector. Detailed proof can be found in    \cite[Sec.\,3]{zhang2022advancing}.

\noindent \textbf{Experiments Results.} 
To implement \ref{eq: prob_pruning},  we adopt two BLO methods: {\IF} (Sec. \ref{sec: IF_UBLO}) and {\GU} (Sec. \ref{sec: GU_UBLO}).
We term the resulting BLO-inspired model pruning approaches as \textbf{BiP-IF} and \textbf{BiP-GU}. For comparison, we also consider two commonly-used non-BLO based pruning methods, the state-of-the-art iterative magnitude pruning (\textbf{IMP})   \cite{frankle2018lottery} and the most efficient one-shot magnitude pruning (\textbf{OMP}) \cite{ma2021sanity}.
We remark that the notable lottery ticket hypothesis (LTH) \cite{frankle2018lottery} stated that IMP is able to identify a trainable sparse subnetwork (known as a `winning ticket'), with test accuracy surprisingly on par or even better than the original model.

\begin{figure*}[tb]
\vspace{-0.5mm}
\centerline{
\begin{tabular}{cccc}
    \includegraphics[width=.23\textwidth,height=!]{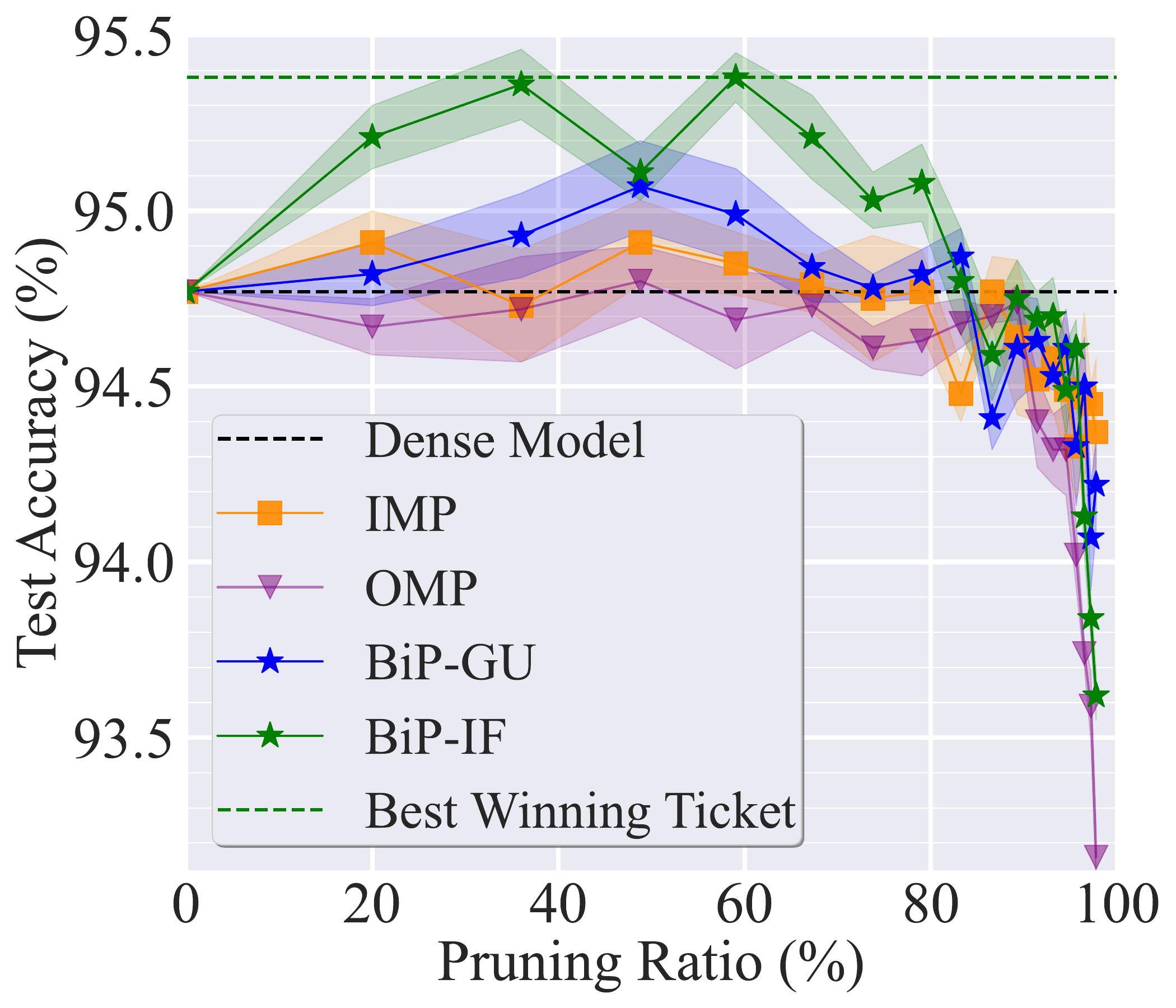} &
    \includegraphics[width=.23\textwidth,height=!]{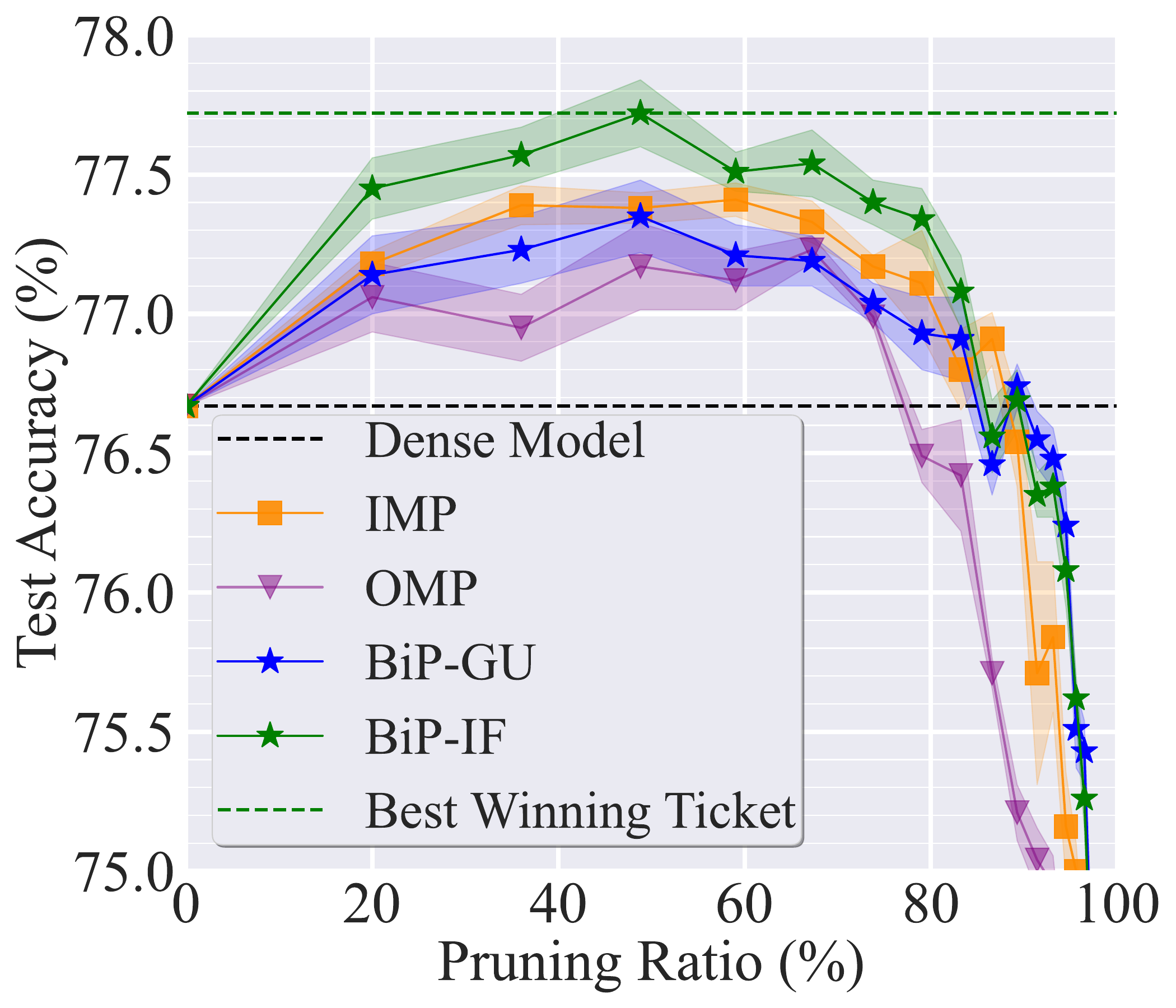} &
    \includegraphics[width=.23\textwidth,height=!]{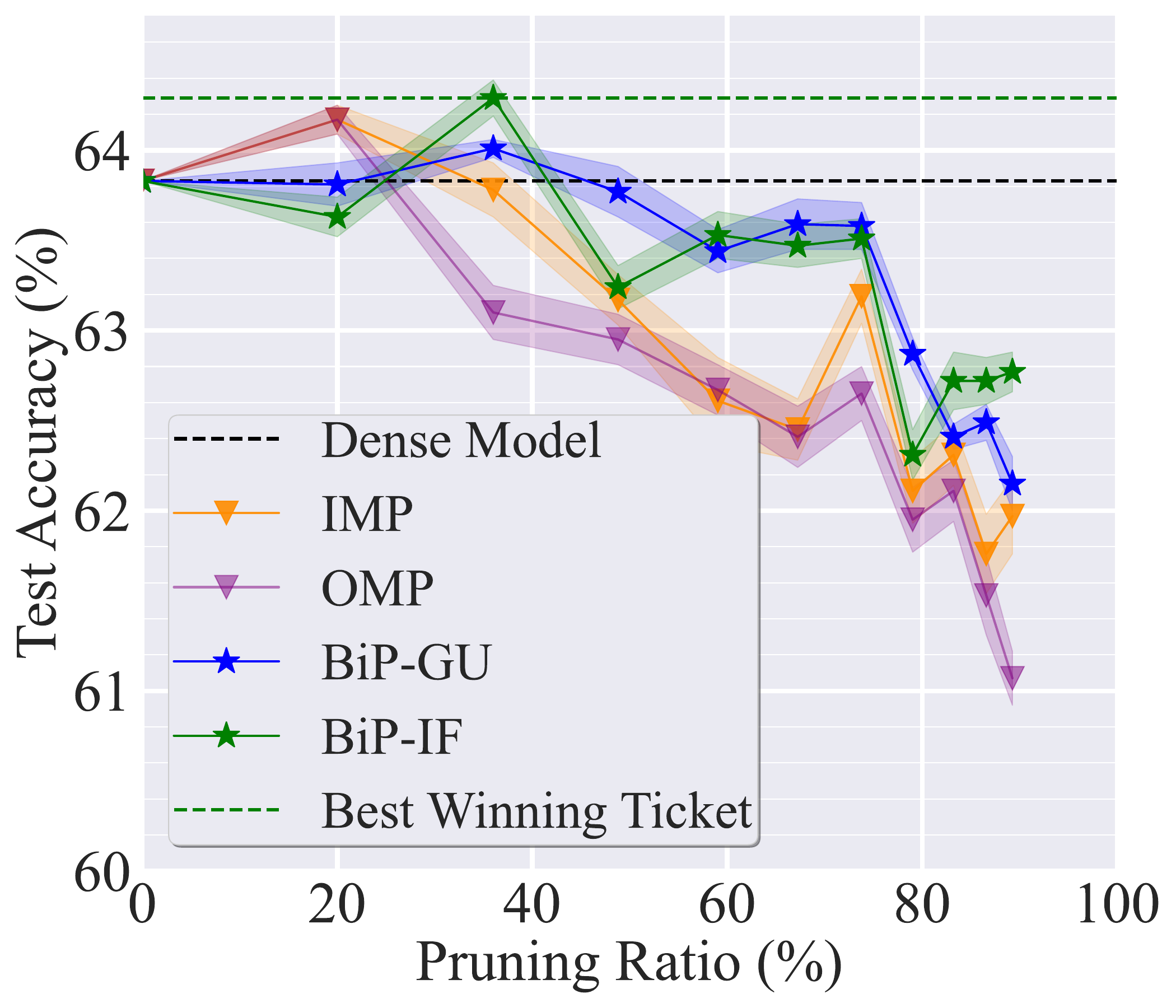} &
    \includegraphics[width=.23\textwidth,height=!]{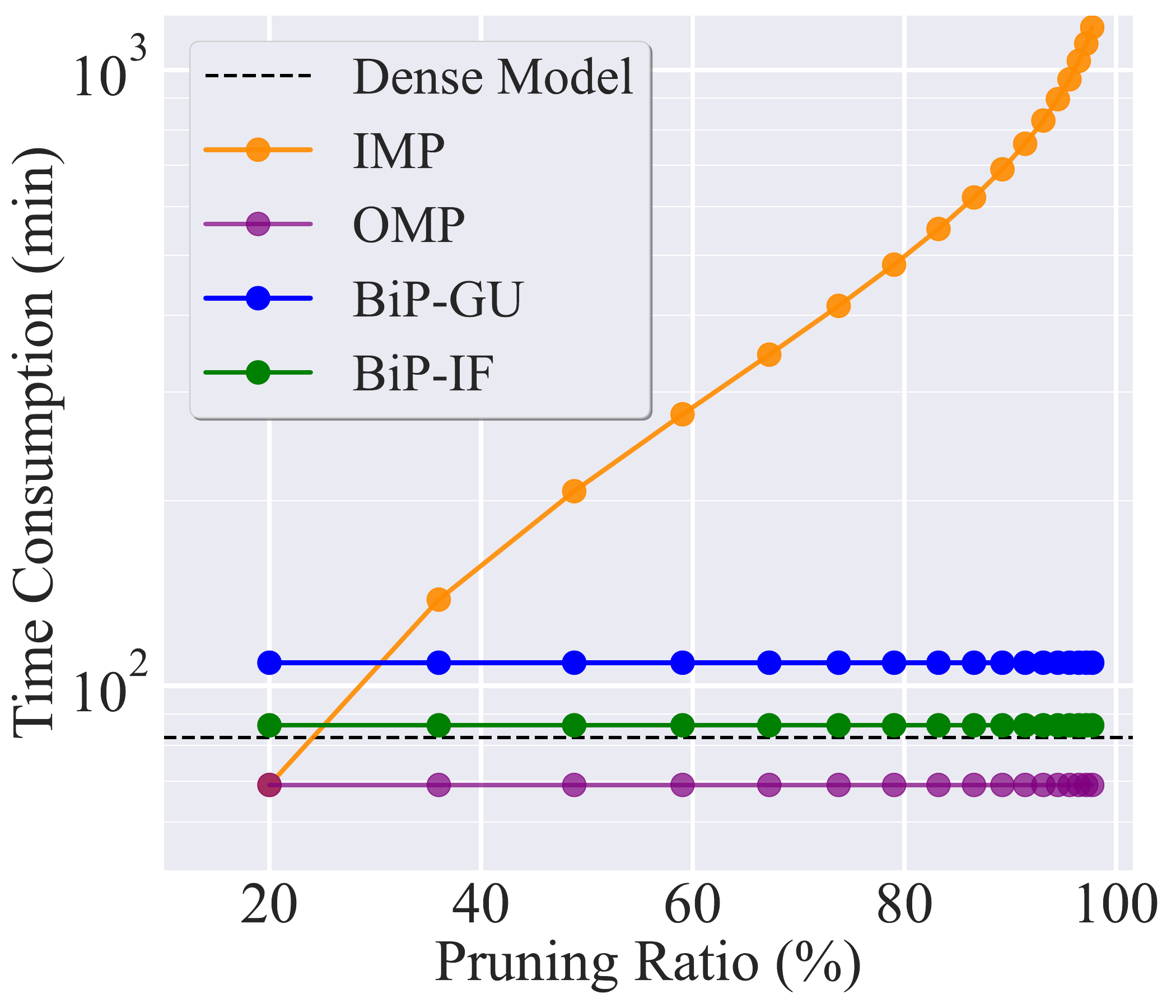}   \\
    \footnotesize{ (a) CIFAR-10 }
    & 
   \footnotesize{ (b) CIFAR-100 }
    & 
  \footnotesize{  (c) Tiny-ImageNet }
    & 
   \footnotesize{ (d) Efficiency comparison }
\end{tabular}}
\caption{{Experiment results of model pruning on different datasets under ResNet-18. (a-c) Pruning trajectory is given by test accuracy (\%) vs. sparsity (\%) under different datasets.
(d) The entire time consumption vs. the pruning ratio.
}}
\label{fig: exp_pruning}
\end{figure*}

\textbf{Fig.\,\ref{fig: exp_pruning}}  illustrates the  pruning accuracy and   the  run-time efficiency of BLO-based pruning methods vs. non-BLO approaches across diverse image classification datasets (including CIFAR-10, CIFAR-100, and Tiny-ImageNet) under ResNet-18.
As we can see, BiP-IF yields the best performance in all the dataset settings evidenced in Fig.\,\ref{fig: exp_pruning}(a)-(c). This also the pruning recipe used in \cite{zhang2022advancing}.
Thanks to the closed-form expression of {\IG} in \eqref{eq: IG_final},   the computation of BiP-IF is also more efficient  than  BiP-GU, as shown in Fig.\,\ref{fig: exp_pruning}(d). In addition, BiP-GU can also provide   competitive pruning accuracy to IMP and takes less computation time than IMP. 
Furthermore, we observe that OMP yields the least computation time but the worst pruning accuracy. This is not surprising since OMP  adopts a non-iterative pruning scheme to find the model's sparse pattern. 

\section{BLO for Invariant Representation Learning}
\label{sec: general_AI}
In this section, we explore the application of BLO in improving the generalization capabilities of ML models. Specifically, we investigate the use of BLO for acquiring training environment-agnostic data representations through invariant risk minimization (\textbf{IRM}) \cite{arjovsky2019invariant}. The goal is to enhance the domain generalization of ML models.

\noindent \textbf{Formulation.}
IRM \cite{arjovsky2019invariant} is 
proposed to acquire invariant data representations and  to enforce invariant predictions against distribution shifts. 
Different from the conventional environment risk minimization (\textbf{ERM})-based training,  IRM yields the BLO-like formulation: 
The upper-level optimization task of IRM is to train a network backbone  to  capture environment-agnostic data representations, and the lower-level optimization task is to find an invariant prediction head (one top of the learned representation network) to produce a global optima to all the training environments. 
Formally, IRM can be cast as

\vspace*{-5mm}
{\small
 \begin{align}
    \displaystyle \minimize_{\boldsymbol \theta} \sum_{i=1}^E \ell_{i}( \boldsymbol \phi^* (\boldsymbol \theta) \circ  \boldsymbol \theta  ) ~~ \st ~~ \boldsymbol \phi^* (\boldsymbol \theta) \in \argmin_{\boldsymbol \phi} \ell_{i} ( \boldsymbol \phi \circ \boldsymbol \theta ),
    \label{eq: irm}
   \tag{IRM}
 \end{align}}%
where $\boldsymbol \phi \circ \boldsymbol \theta$ denotes the representation-acquisition model $\boldsymbol \theta$ and the predictor $\boldsymbol \phi$, $\ell_{i}$ is the  training loss associated with the $i$th training environment, and $E$ is the total number of training environments. The rationale behind \eqref{eq: irm} is that given the invariant  representation extractor $\boldsymbol \theta$, there exists an invariant predictor $ \boldsymbol \phi^* (\boldsymbol \theta)$ which is optimal across all the training environments.

\noindent \textbf{Methods.} 
Solving problem \eqref{eq: irm} is highly non-trivial since the lower-level solution $\boldsymbol \phi(\btheta)$ should be \textit{universal} across all $E$ training environments. To circumvent this difficulty, \ref{eq: irm} is typically relaxed to a \textit{single-level} optimization problem, known as IRMv1 \cite{arjovsky2019invariant}:

\vspace*{-5mm}
{
\small
\begin{align}
    \minimize_{\btheta} \sum_{i=1}^E [\ell_i(\btheta) + \gamma \|\nabla_{w|w=1.0} \ell_i (w \circ \btheta)\|_2^2]
\end{align}
}%
where $\gamma > 0$ is a regularization parameter and $\nabla_{w|w=1.0} \ell_i (w \odot \btheta)$ denotes the gradient of $\ell_i$ \textit{w.r.t.} w, computed at $w=1.0$.
In the formulation above, the identity mapping $w = 1.0$ is adopted, symbolizing a basic ``imaginary'' classification head. Meanwhile, $\btheta$ corresponds to the combination of the representation extractor and the actual invariant predictor.
However, such a simplification is restricted to linear invariant prediction and penalizes the deviation of individual environment losses from stationarity to approach the lower-level optimality in \ref{eq: irm}.
Despite the practical simplicity of IRMv1, it may fail to achieve the desired invariance \cite{chen2022pareto, kamath2021does}.

Beyond IRMv1, a consensus-constrained BLO method is developed in \cite{zhang2023what} to solve problem \eqref{eq: irm}.
The key idea is to introduce $E$ auxiliary predictors  $\{ \boldsymbol \phi_i \}$ and
explicitly enforce prediction invariance by   infusing a \textit{consensus prediction constraint} $\mathcal C = \{ \{ \bphi_i\} | \bphi_1 = \ldots = \bphi_E \}$ to the lower-level problem of \eqref{eq: irm}, and   promote the per-environment stationarity in its upper-level problem. 
This modifies \eqref{eq: irm} to an ordinary BLO problem:

\vspace*{-5mm}
{\small \begin{align}
\begin{array}{ll}
\displaystyle \minimize_{\boldsymbol \theta}     & \sum_{i=1}^E [ \ell_{i}( \boldsymbol \phi_i^* (\boldsymbol \theta) \circ  \boldsymbol \theta  ) + \gamma \| \nabla_{\boldsymbol \phi_i} \ell_i ( \boldsymbol \phi_i^* (\boldsymbol \theta) \circ  \boldsymbol \theta  ) \|_2^2 ]\\
\st & \{  \boldsymbol \phi_i^* (\boldsymbol \theta) \}_{i=1}^E = \displaystyle \argmin_{ \{ \bphi_i\} \in \mathcal C }  \sum_{i=1}^E \ell_{i} ( \boldsymbol \phi_i \circ \boldsymbol \theta ), ~\forall i \in [E],
\end{array} 
\label{eq: irm_game}
\tag{IRM-BLO}
\end{align}}%
where $\gamma > 0$ is a regularization parameter, and $[E]$ denotes the integer set $\{1,2,\ldots, E\}$.
The advantage of converting  \eqref{eq: irm} into  the consensus-constrained  \eqref{eq: irm_game} is that projection onto the consensus constraint yields a closed-form solution, \textit{i.e.},  
$ \mathcal{P}_{\mathcal C} (\mathbf a) = \argmin_{\{\boldsymbol \phi_i \} \in \mathcal C} \sum_{i=1}^E \| \boldsymbol \phi_i - \mathbf a_i \|_2^2 = \frac{1}{E} \sum_i \mathbf a_i$, where $\mathcal{P}_{\mathcal C}(\mathbf a)$  denotes the projection  operation to project the point $\mathbf a$ onto the constraint $\mathcal C$.
It has been shown in \cite{zhang2023what} that problem \eqref{eq: irm_game} can be effectively solved using the {\GU} approach, which approximates each individual lower-level solution using $K$-step GD unrolling together with the consensus projection. Thus, the lower-level solution becomes

\vspace*{-5mm}
{\small \begin{align}\label{eq: inner_GD_IRM}
\forall i, ~~\boldsymbol \phi_i^* (\boldsymbol \theta) \approx \frac{1}{E} \sum_{i=1}^E \boldsymbol \phi_i^{(K)}, ~~
\boldsymbol \phi_i^{(k)} = \boldsymbol \phi_i^{(k-1)}- \beta \nabla_{\boldsymbol \phi_i}
\ell_i(\boldsymbol \phi_i^{(k-1)} \circ \boldsymbol \theta),
\text{ for }
k \in [K],
\end{align}
}%
where $\beta > 0$ is the lower-level learning rate. Based on the above, automatic differentiation (AD) can be called to compute the implicit gradient from  $\boldsymbol \phi_i^{(K)} (\boldsymbol \theta)$ to the variable $\btheta$ in the {\GU} process.

\begin{table}[htb]
\centering
\caption{{Performance of different  IRM training methods. The best performance per evaluation metric is highlighted in \textbf{bold} and the performance of the BLO-enabled method is marked in \colorbox{Gray}{gray}. 
}}
\label{tab: irm_overview}
\resizebox{.5\textwidth}{!}{%
\begin{tabular}{c|c|cc|cc}
\toprule[1pt]
\midrule
\multirow{2}{*}{\textbf{Method}} 
& \multirow{2}{*}{\textbf{\begin{tabular}[c]{@{}c@{}}BLO\\(Solver)\end{tabular}}} 
& \multicolumn{2}{c|}{\textbf{Colored-MNIST}} & \multicolumn{2}{c}{\textbf{Colored-FashionMNIST}}\\ 
& 
&\textbf{Avg. Acc.} & \textbf{Acc. Gap} & \textbf{Avg. Acc.} & \textbf{Acc. Gap} \\
\midrule
ERM
& N/A
& 49.19\footnotesize{$\pm1.89$} 
& 90.72\footnotesize{$\pm2.08$}
& 49.77\footnotesize{$\pm1.71$} 
& 88.62\footnotesize{$\pm2.49$}
\\
IRM-v1 \cite{arjovsky2019invariant}
& N/A
& 68.33\footnotesize{$\pm0.31$}     
& 2.04\footnotesize{$\pm0.05$}  
& 68.76\footnotesize{$\pm0.31$} 
& 1.45\footnotesize{$\pm0.09$}
\\
IRM-Game \cite{ahuja2020invariant}
& N/A
& 67.73\footnotesize{$\pm0.24$} 
& 1.67\footnotesize{$\pm0.14$} 
& 67.49\footnotesize{$\pm0.32$} 
& 1.82\footnotesize{$\pm0.13$}
\\
\midrule
\rowcolor{Gray}
IRM-BLO \cite{zhang2023what}
& GU
& \textbf{69.47}\footnotesize{$\pm0.24$} 
& \textbf{1.04}\footnotesize{$\pm0.07$} 
& \textbf{69.43}\footnotesize{$\pm0.21$} 
& \textbf{1.14}\footnotesize{$\pm0.11$} 
\\
\midrule
\bottomrule[1pt]
\end{tabular}%
}
\end{table}

\noindent \textbf{Experiment Results.} 
We evaluate the  performance of \ref{eq: irm_game} with two commonly-used  image classification datasets, Colored-MNIST \cite{arjovsky2019invariant} and Colored-FashionMNIST \cite{ahuja2020invariant}, where  spurious  correlation is imposed in the datasets which makes the conventional ERM training ineffective.
To capture both the accuracy and the variance of invariant predictions across multiple testing environments, the \textit{average accuracy} and the \textit{accuracy gap} (the difference between best and worst-case accuracy) are measured for IRM methods.
\textbf{Tab.\,\ref{tab: irm_overview}} presents the resulting performance of \ref{eq: irm_game} and compares it with that of ERM and two IRM baselines, IRMv1 \cite{arjovsky2019invariant}and IRM-Game \cite{ahuja2020invariant}.
Note that all IRM variants outperform ERM, which justifies the importance of IRM training to improve model generalization across diverse environments.
Within the IRM training family, IRM-BLO outperforms others by achieving the highest average accuracy and the smallest accuracy gap across both datasets. This superior performance underscores the value of BLO compared to the sub-optimal design IRM-v1.

\section{Discussion}

BLO is a challenging but rapidly developing subject.
Despite recent progresses discussed in this article, significant work is yet to be done to address various challenges ranging from developing scalable algorithms, to extending its applicability to a wider range of problems. Below we highlight several worthwhile future directions.
    
$\bullet$ \textbf{BLO algorithms.} 
\textbf{First}, the development and analysis of BLO algorithms for more general BLO problems, including those with complex lower-level constraints (\textit{e.g.}, non-linear constraints), require additional exploration. The current focus has predominantly been on problems with linear constraints, and extending the BLO framework to handle non-linear constraints is an important and challenging task; some recent work towards this direction in the context of the min-max problems can be found in \cite{Tsaknakis23}. Additionally, exploring scenarios with coupled constraints between lower and upper levels, such as resource sharing among adversarial and normal agents, presents a complex area that is under-explored.
\textbf{Second}, BLO formulations with non-singleton lower-level solutions, such as \eqref{eq: Optimistic_BLO}, lack theoretically grounded, scalable, and easy-to-implement algorithms. This aspect of BLO has received less attention from the community, making it an open topic for future investigation. Developing efficient algorithms that can handle non-singleton lower-level solutions and provide convergence guarantees is a key research direction.
\textbf{Third}, beyond the scope of bi-level optimization, exploring problems involving more than two levels, such as dataset pruning for transfer learning, represents an exciting frontier. These multi-level problems introduce additional complexity and challenges, requiring the design of novel algorithms to tackle the inherent hierarchical structure effectively. \textbf{Finally}, large-scale and distributed availability of data demands the development of decentralized and federated algorithms to solve these complex BLO problems which also presents a compelling research direction for future exploration.
    
$\bullet$ \textbf{BLO theories.} 
\textbf{First}, while significant progress has been made in establishing theoretical guarantees for solving the basic BLO problem \eqref{eq: prob_basic_BLO}, more attention is needed on exploring practical settings including (coupled) lower-level constraints, non-convex lower-level problems, and/or black-box settings where one may not have access to upper/lower-level parameters. These scenarios present unique challenges and complexities, making it difficult to analyze the problem and derive theoretical guarantees.  Investigating the convergence properties and establishing theoretical foundations for solving BLO problems under these practical settings is an important avenue for future research.
\textbf{Second}, when datasets become   massive, it is crucial to develop  theoretically-grounded BLO algorithms that can adhere to practical requirements. Therefore, developing theoretical frameworks and analyzing the convergence properties of algorithms for solving large-scale BLO problems under realistic assumptions is also an important  research topic. \textbf{Third}, with the discovery of the phenomenon of double descent \cite{belkin2019reconciling}, theoretical analysis of standard ML algorithms on overparameterized neural networks has received significant attention from the research community. Theoretical investigation of BLO algorithms for such overparameterized problems is certainly an interesting research direction. \textbf{Finally}, developing a thorough theoretical understanding of distributed/federated algorithms for solving complex BLO problems will certainly be at the forefront of future BLO research.
    
$\bullet$  \textbf{BLO applications.} 
\textbf{First}, in the context of Mixture-of-Experts (MoE) training, there is a complex interplay between the training of the gating network that selects experts and the training of the expert-oriented pathways used for final predictions. Exploring BLO techniques to effectively optimize the coupling between these two training processes in MoE can lead to improved performance and better utilization of emerging ML models like MoE.
\textbf{Second}, prompt learning, a key technique used in today's foundation models, involves a crucial coupling between prompt pattern learning and label/feature mapping optimization \cite{chen2023understanding}. Leveraging BLO methods to model and optimize the interactions between prompt pattern learning and label/feature mapping can enhance the learning process and enable accurate and robust  prompt generations.
\textbf{Third}, BLO is highly applicable in (inverse) reinforcement learning. For instance, the Actor/Critic Algorithm can be formulated as a BLO problem, with separate agents evaluating and optimizing the policy. In {\it inverse} reinforcement learning, the tasks involve inferring the agents' reward function and finding the optimal policy based on it. Applying BLO frameworks to these scenarios offers potential for novel insights and improved efficiency. 

In summary, the interplay between the theoretical underpinnings of BLO and its practical applications promises a fertile ground for future exploration and innovation, pushing the boundaries of optimization theory and applications in SP and ML.

\section*{Acknowledgement}
The work of Y. Zhang, Y. Yao, and S. Liu was supported by the NSF Grant IIS-2207052.

\clearpage

\bibliographystyle{IEEEbib}
\bibliography{refs}

\end{document}